%% file: main.tex
\documentclass[acmsmall]{acmart}

\AtBeginDocument{%
  }

\setcopyright{acmcopyright}
\copyrightyear{2018}
\acmYear{2018}
\acmDOI{XXXXXXX.XXXXXXX}

\acmJournal{JACM}
\acmVolume{37}
\acmNumber{4}
\acmArticle{111}
\acmMonth{8}

\usepackage{latexsym}
\usepackage{amsmath}
\usepackage{url}
\usepackage{amsmath,nccmath,amsfonts}
\usepackage{algorithm}
\usepackage{algorithmic}
\usepackage{xcolor}
\usepackage{graphicx}
\usepackage{subfigure}
\usepackage{wrapfig}
\usepackage{relsize}
\usepackage{caption}

\usepackage{bbm}

\usepackage{enumitem}

\usepackage{array}
\newcolumntype{P}[1]{>{\centering\arraybackslash}p{#1}}

\usepackage{multirow}

\newcommand{\tianxiang}[1]{{\color{black}#1}}
\newcommand{\revision}[1]{{\color{black}#1}}

\begin{document}

\setcopyright{acmlicensed}
\acmJournal{TIST}
\acmYear{2023} \acmVolume{1} \acmNumber{1} \acmArticle{1} \acmMonth{1} \acmPrice{15.00}\acmDOI{10.1145/3616542}

\title{Faithful and Consistent Graph Neural Network Explanations with Rationale Alignment}


\author{Tianxiang Zhao}
\affiliation{%
  \institution{The Pennsylvania State University}
  \city{State College}
  \country{USA}}
\email{tkz5084@psu.edu}

\author{Dongsheng Luo}
\affiliation{%
  \institution{Florida International University}
  \city{Miami}
  \country{USA}}
\email{dluo@fiu.edu}

\author{Xiang Zhang}
\affiliation{%
  \institution{The Pennsylvania State University}
  \city{State College}
  \country{USA}}
\email{xzz89@psu.edu}

\author{Suhang Wang}
\affiliation{%
  \institution{The Pennsylvania State University}
  \city{State College}
  \country{USA}}
\email{szw494@psu.edu}

\renewcommand{\shortauthors}{Zhao et al.}

\begin{abstract}
Uncovering rationales behind predictions of graph neural networks (GNNs) has received increasing attention over recent years. Instance-level GNN explanation aims to discover critical input elements, like nodes or edges, that the target GNN relies upon for making predictions. 
Though various algorithms are proposed, most of them formalize this task by searching the minimal subgraph which can preserve original predictions. However, an inductive bias is deep-rooted in this framework: several subgraphs can result in the same or similar outputs as the original graphs. Consequently, they have the danger of providing spurious explanations and failing to provide consistent explanations. Applying them to explain weakly-performed GNNs would further amplify these issues. To address this problem, we theoretically examine the predictions of GNNs from the causality perspective. Two typical reasons for spurious explanations are identified: confounding effect of latent variables like distribution shift, and causal factors distinct from the original input. Observing that both confounding effects and diverse causal rationales are encoded in internal representations, \tianxiang{we propose a new explanation framework with an auxiliary alignment loss, which is theoretically proven to be optimizing a more faithful explanation objective intrinsically. Concretely for this alignment loss, a set of different perspectives are explored: anchor-based alignment, distributional alignment based on Gaussian mixture models, mutual-information-based alignment, etc. A comprehensive study is conducted both on the effectiveness of this new framework in terms of explanation faithfulness/consistency and on the advantages of these variants. For our codes, please refer to the following URL link: \url{https://github.com/TianxiangZhao/GraphNNExplanation}.}
\end{abstract}



\received{13 March 2023}

\maketitle

\input{math_command.tex}

\input{introduction.tex}

\input{related_work.tex}

\input{problem_setting.tex}

\input{methodology.tex}

\input{experiment.tex}

\input{conclusion.tex}

\begin{acks}

This material is based upon work supported by, or in part by, the National Science Foundation under grants number IIS-1707548 and IIS-1909702, the Army Research Office under grant number W911NF21-1-0198, and DHS CINA under grant number E205949D. The findings and conclusions in this paper do not necessarily reflect the view of the funding agency.
\end{acks}

\bibliographystyle{unsrt}
\bibliography{bib1}


\end{document}

%% file: math_command.tex

\newcommand{\figleft}{{\em (Left)}}
\newcommand{\figcenter}{{\em (Center)}}
\newcommand{\figright}{{\em (Right)}}
\newcommand{\figtop}{{\em (Top)}}
\newcommand{\figbottom}{{\em (Bottom)}}
\newcommand{\captiona}{{\em (a)}}
\newcommand{\captionb}{{\em (b)}}
\newcommand{\captionc}{{\em (c)}}
\newcommand{\captiond}{{\em (d)}}

\newcommand{\newterm}[1]{{\bf #1}}

\def\figref#1{figure~\ref{#1}}
\def\Figref#1{Figure~\ref{#1}}
\def\twofigref#1#2{figures \ref{#1} and \ref{#2}}
\def\quadfigref#1#2#3#4{figures \ref{#1}, \ref{#2}, \ref{#3} and \ref{#4}}
\def\secref#1{section~\ref{#1}}
\def\Secref#1{Section~\ref{#1}}
\def\twosecrefs#1#2{sections \ref{#1} and \ref{#2}}
\def\secrefs#1#2#3{sections \ref{#1}, \ref{#2} and \ref{#3}}
\def\eqref#1{equation~\ref{#1}}
\def\Eqref#1{Equation~\ref{#1}}
\def\plaineqref#1{\ref{#1}}
\def\chapref#1{chapter~\ref{#1}}
\def\Chapref#1{Chapter~\ref{#1}}
\def\rangechapref#1#2{chapters\ref{#1}--\ref{#2}}
\def\algref#1{algorithm~\ref{#1}}
\def\Algref#1{Algorithm~\ref{#1}}
\def\twoalgref#1#2{algorithms \ref{#1} and \ref{#2}}
\def\Twoalgref#1#2{Algorithms \ref{#1} and \ref{#2}}
\def\partref#1{part~\ref{#1}}
\def\Partref#1{Part~\ref{#1}}
\def\twopartref#1#2{parts \ref{#1} and \ref{#2}}

\def\ceil#1{\lceil #1 \rceil}
\def\floor#1{\lfloor #1 \rfloor}
\def\1{\bm{1}}
\newcommand{\train}{\mathcal{D}}
\newcommand{\valid}{\mathcal{D_{\mathrm{valid}}}}
\newcommand{\test}{\mathcal{D_{\mathrm{test}}}}

\def\eps{{\epsilon}}

\def\reta{{\textnormal{$\eta$}}}
\def\ra{{\textnormal{a}}}
\def\rb{{\textnormal{b}}}
\def\rc{{\textnormal{c}}}
\def\rd{{\textnormal{d}}}
\def\re{{\textnormal{e}}}
\def\rf{{\textnormal{f}}}
\def\rg{{\textnormal{g}}}
\def\rh{{\textnormal{h}}}
\def\ri{{\textnormal{i}}}
\def\rj{{\textnormal{j}}}
\def\rk{{\textnormal{k}}}
\def\rl{{\textnormal{l}}}
\def\rn{{\textnormal{n}}}
\def\ro{{\textnormal{o}}}
\def\rp{{\textnormal{p}}}
\def\rq{{\textnormal{q}}}
\def\rr{{\textnormal{r}}}
\def\rs{{\textnormal{s}}}
\def\rt{{\textnormal{t}}}
\def\ru{{\textnormal{u}}}
\def\rv{{\textnormal{v}}}
\def\rw{{\textnormal{w}}}
\def\rx{{\textnormal{x}}}
\def\ry{{\textnormal{y}}}
\def\rz{{\textnormal{z}}}

\def\rvepsilon{{\mathbf{\epsilon}}}
\def\rvtheta{{\mathbf{\theta}}}
\def\rva{{\mathbf{a}}}
\def\rvb{{\mathbf{b}}}
\def\rvc{{\mathbf{c}}}
\def\rvd{{\mathbf{d}}}
\def\rve{{\mathbf{e}}}
\def\rvf{{\mathbf{f}}}
\def\rvg{{\mathbf{g}}}
\def\rvh{{\mathbf{h}}}
\def\rvu{{\mathbf{i}}}
\def\rvj{{\mathbf{j}}}
\def\rvk{{\mathbf{k}}}
\def\rvl{{\mathbf{l}}}
\def\rvm{{\mathbf{m}}}
\def\rvn{{\mathbf{n}}}
\def\rvo{{\mathbf{o}}}
\def\rvp{{\mathbf{p}}}
\def\rvq{{\mathbf{q}}}
\def\rvr{{\mathbf{r}}}
\def\rvs{{\mathbf{s}}}
\def\rvt{{\mathbf{t}}}
\def\rvu{{\mathbf{u}}}
\def\rvv{{\mathbf{v}}}
\def\rvw{{\mathbf{w}}}
\def\rvx{{\mathbf{x}}}
\def\rvy{{\mathbf{y}}}
\def\rvz{{\mathbf{z}}}

\def\erva{{\textnormal{a}}}
\def\ervb{{\textnormal{b}}}
\def\ervc{{\textnormal{c}}}
\def\ervd{{\textnormal{d}}}
\def\erve{{\textnormal{e}}}
\def\ervf{{\textnormal{f}}}
\def\ervg{{\textnormal{g}}}
\def\ervh{{\textnormal{h}}}
\def\ervi{{\textnormal{i}}}
\def\ervj{{\textnormal{j}}}
\def\ervk{{\textnormal{k}}}
\def\ervl{{\textnormal{l}}}
\def\ervm{{\textnormal{m}}}
\def\ervn{{\textnormal{n}}}
\def\ervo{{\textnormal{o}}}
\def\ervp{{\textnormal{p}}}
\def\ervq{{\textnormal{q}}}
\def\ervr{{\textnormal{r}}}
\def\ervs{{\textnormal{s}}}
\def\ervt{{\textnormal{t}}}
\def\ervu{{\textnormal{u}}}
\def\ervv{{\textnormal{v}}}
\def\ervw{{\textnormal{w}}}
\def\ervx{{\textnormal{x}}}
\def\ervy{{\textnormal{y}}}
\def\ervz{{\textnormal{z}}}

\def\rmA{{\mathbf{A}}}
\def\rmB{{\mathbf{B}}}
\def\rmC{{\mathbf{C}}}
\def\rmD{{\mathbf{D}}}
\def\rmE{{\mathbf{E}}}
\def\rmF{{\mathbf{F}}}
\def\rmG{{\mathbf{G}}}
\def\rmH{{\mathbf{H}}}
\def\rmI{{\mathbf{I}}}
\def\rmJ{{\mathbf{J}}}
\def\rmK{{\mathbf{K}}}
\def\rmL{{\mathbf{L}}}
\def\rmM{{\mathbf{M}}}
\def\rmN{{\mathbf{N}}}
\def\rmO{{\mathbf{O}}}
\def\rmP{{\mathbf{P}}}
\def\rmQ{{\mathbf{Q}}}
\def\rmR{{\mathbf{R}}}
\def\rmS{{\mathbf{S}}}
\def\rmT{{\mathbf{T}}}
\def\rmU{{\mathbf{U}}}
\def\rmV{{\mathbf{V}}}
\def\rmW{{\mathbf{W}}}
\def\rmX{{\mathbf{X}}}
\def\rmY{{\mathbf{Y}}}
\def\rmZ{{\mathbf{Z}}}

\def\ermA{{\textnormal{A}}}
\def\ermB{{\textnormal{B}}}
\def\ermC{{\textnormal{C}}}
\def\ermD{{\textnormal{D}}}
\def\ermE{{\textnormal{E}}}
\def\ermF{{\textnormal{F}}}
\def\ermG{{\textnormal{G}}}
\def\ermH{{\textnormal{H}}}
\def\ermI{{\textnormal{I}}}
\def\ermJ{{\textnormal{J}}}
\def\ermK{{\textnormal{K}}}
\def\ermL{{\textnormal{L}}}
\def\ermM{{\textnormal{M}}}
\def\ermN{{\textnormal{N}}}
\def\ermO{{\textnormal{O}}}
\def\ermP{{\textnormal{P}}}
\def\ermQ{{\textnormal{Q}}}
\def\ermR{{\textnormal{R}}}
\def\ermS{{\textnormal{S}}}
\def\ermT{{\textnormal{T}}}
\def\ermU{{\textnormal{U}}}
\def\ermV{{\textnormal{V}}}
\def\ermW{{\textnormal{W}}}
\def\ermX{{\textnormal{X}}}
\def\ermY{{\textnormal{Y}}}
\def\ermZ{{\textnormal{Z}}}

\def\vzero{{\bm{0}}}
\def\vone{{\bm{1}}}
\def\vmu{{\bm{\mu}}}
\def\vtheta{{\bm{\theta}}}
\def\va{{\bm{a}}}
\def\vb{{\bm{b}}}
\def\vc{{\bm{c}}}
\def\vd{{\bm{d}}}
\def\ve{{\bm{e}}}
\def\vf{{\bm{f}}}
\def\vg{{\bm{g}}}
\def\vh{{\bm{h}}}
\def\vi{{\bm{i}}}
\def\vj{{\bm{j}}}
\def\vk{{\bm{k}}}
\def\vl{{\bm{l}}}
\def\vm{{\bm{m}}}
\def\vn{{\bm{n}}}
\def\vo{{\bm{o}}}
\def\vp{{\bm{p}}}
\def\vq{{\bm{q}}}
\def\vr{{\bm{r}}}
\def\vs{{\bm{s}}}
\def\vt{{\bm{t}}}
\def\vu{{\bm{u}}}
\def\vv{{\bm{v}}}
\def\vw{{\bm{w}}}
\def\vx{{\bm{x}}}
\def\vy{{\bm{y}}}
\def\vz{{\bm{z}}}

\def\evalpha{{\alpha}}
\def\evbeta{{\beta}}
\def\evepsilon{{\epsilon}}
\def\evlambda{{\lambda}}
\def\evomega{{\omega}}
\def\evmu{{\mu}}
\def\evpsi{{\psi}}
\def\evsigma{{\sigma}}
\def\evtheta{{\theta}}
\def\eva{{a}}
\def\evb{{b}}
\def\evc{{c}}
\def\evd{{d}}
\def\eve{{e}}
\def\evf{{f}}
\def\evg{{g}}
\def\evh{{h}}
\def\evi{{i}}
\def\evj{{j}}
\def\evk{{k}}
\def\evl{{l}}
\def\evm{{m}}
\def\evn{{n}}
\def\evo{{o}}
\def\evp{{p}}
\def\evq{{q}}
\def\evr{{r}}
\def\evs{{s}}
\def\evt{{t}}
\def\evu{{u}}
\def\evv{{v}}
\def\evw{{w}}
\def\evx{{x}}
\def\evy{{y}}
\def\evz{{z}}

\def\mA{{\bm{A}}}
\def\mB{{\bm{B}}}
\def\mC{{\bm{C}}}
\def\mD{{\bm{D}}}
\def\mE{{\bm{E}}}
\def\mF{{\bm{F}}}
\def\mG{{\bm{G}}}
\def\mH{{\bm{H}}}
\def\mI{{\bm{I}}}
\def\mJ{{\bm{J}}}
\def\mK{{\bm{K}}}
\def\mL{{\bm{L}}}
\def\mM{{\bm{M}}}
\def\mN{{\bm{N}}}
\def\mO{{\bm{O}}}
\def\mP{{\bm{P}}}
\def\mQ{{\bm{Q}}}
\def\mR{{\bm{R}}}
\def\mS{{\bm{S}}}
\def\mT{{\bm{T}}}
\def\mU{{\bm{U}}}
\def\mV{{\bm{V}}}
\def\mW{{\bm{W}}}
\def\mX{{\bm{X}}}
\def\mY{{\bm{Y}}}
\def\mZ{{\bm{Z}}}
\def\mBeta{{\bm{\beta}}}
\def\mPhi{{\bm{\Phi}}}
\def\mLambda{{\bm{\Lambda}}}
\def\mSigma{{\bm{\Sigma}}}


\newcommand{\tens}[1]{\bm{\mathsfit{#1}}}
\def\tA{{\tens{A}}}
\def\tB{{\tens{B}}}
\def\tC{{\tens{C}}}
\def\tD{{\tens{D}}}
\def\tE{{\tens{E}}}
\def\tF{{\tens{F}}}
\def\tG{{\tens{G}}}
\def\tH{{\tens{H}}}
\def\tI{{\tens{I}}}
\def\tJ{{\tens{J}}}
\def\tK{{\tens{K}}}
\def\tL{{\tens{L}}}
\def\tM{{\tens{M}}}
\def\tN{{\tens{N}}}
\def\tO{{\tens{O}}}
\def\tP{{\tens{P}}}
\def\tQ{{\tens{Q}}}
\def\tR{{\tens{R}}}
\def\tS{{\tens{S}}}
\def\tT{{\tens{T}}}
\def\tU{{\tens{U}}}
\def\tV{{\tens{V}}}
\def\tW{{\tens{W}}}
\def\tX{{\tens{X}}}
\def\tY{{\tens{Y}}}
\def\tZ{{\tens{Z}}}

\def\gA{{\mathcal{A}}}
\def\gB{{\mathcal{B}}}
\def\gC{{\mathcal{C}}}
\def\gD{{\mathcal{D}}}
\def\gE{{\mathcal{E}}}
\def\gF{{\mathcal{F}}}
\def\gG{{\mathcal{G}}}
\def\gH{{\mathcal{H}}}
\def\gI{{\mathcal{I}}}
\def\gJ{{\mathcal{J}}}
\def\gK{{\mathcal{K}}}
\def\gL{{\mathcal{L}}}
\def\gM{{\mathcal{M}}}
\def\gN{{\mathcal{N}}}
\def\gO{{\mathcal{O}}}
\def\gP{{\mathcal{P}}}
\def\gQ{{\mathcal{Q}}}
\def\gR{{\mathcal{R}}}
\def\gS{{\mathcal{S}}}
\def\gT{{\mathcal{T}}}
\def\gU{{\mathcal{U}}}
\def\gV{{\mathcal{V}}}
\def\gW{{\mathcal{W}}}
\def\gX{{\mathcal{X}}}
\def\gY{{\mathcal{Y}}}
\def\gZ{{\mathcal{Z}}}

\def\sA{{\mathbb{A}}}
\def\sB{{\mathbb{B}}}
\def\sC{{\mathbb{C}}}
\def\sD{{\mathbb{D}}}
\def\sF{{\mathbb{F}}}
\def\sG{{\mathbb{G}}}
\def\sH{{\mathbb{H}}}
\def\sI{{\mathbb{I}}}
\def\sJ{{\mathbb{J}}}
\def\sK{{\mathbb{K}}}
\def\sL{{\mathbb{L}}}
\def\sM{{\mathbb{M}}}
\def\sN{{\mathbb{N}}}
\def\sO{{\mathbb{O}}}
\def\sP{{\mathbb{P}}}
\def\sQ{{\mathbb{Q}}}
\def\sR{{\mathbb{R}}}
\def\sS{{\mathbb{S}}}
\def\sT{{\mathbb{T}}}
\def\sU{{\mathbb{U}}}
\def\sV{{\mathbb{V}}}
\def\sW{{\mathbb{W}}}
\def\sX{{\mathbb{X}}}
\def\sY{{\mathbb{Y}}}
\def\sZ{{\mathbb{Z}}}

\def\emLambda{{\Lambda}}
\def\emA{{A}}
\def\emB{{B}}
\def\emC{{C}}
\def\emD{{D}}
\def\emE{{E}}
\def\emF{{F}}
\def\emG{{G}}
\def\emH{{H}}
\def\emI{{I}}
\def\emJ{{J}}
\def\emK{{K}}
\def\emL{{L}}
\def\emM{{M}}
\def\emN{{N}}
\def\emO{{O}}
\def\emP{{P}}
\def\emQ{{Q}}
\def\emR{{R}}
\def\emS{{S}}
\def\emT{{T}}
\def\emU{{U}}
\def\emV{{V}}
\def\emW{{W}}
\def\emX{{X}}
\def\emY{{Y}}
\def\emZ{{Z}}
\def\emSigma{{\Sigma}}

\newcommand{\etens}[1]{\mathsfit{#1}}
\def\etLambda{{\etens{\Lambda}}}
\def\etA{{\etens{A}}}
\def\etB{{\etens{B}}}
\def\etC{{\etens{C}}}
\def\etD{{\etens{D}}}
\def\etE{{\etens{E}}}
\def\etF{{\etens{F}}}
\def\etG{{\etens{G}}}
\def\etH{{\etens{H}}}
\def\etI{{\etens{I}}}
\def\etJ{{\etens{J}}}
\def\etK{{\etens{K}}}
\def\etL{{\etens{L}}}
\def\etM{{\etens{M}}}
\def\etN{{\etens{N}}}
\def\etO{{\etens{O}}}
\def\etP{{\etens{P}}}
\def\etQ{{\etens{Q}}}
\def\etR{{\etens{R}}}
\def\etS{{\etens{S}}}
\def\etT{{\etens{T}}}
\def\etU{{\etens{U}}}
\def\etV{{\etens{V}}}
\def\etW{{\etens{W}}}
\def\etX{{\etens{X}}}
\def\etY{{\etens{Y}}}
\def\etZ{{\etens{Z}}}

\newcommand{\pdata}{p_{\rm{data}}}
\newcommand{\ptrain}{\hat{p}_{\rm{data}}}
\newcommand{\Ptrain}{\hat{P}_{\rm{data}}}
\newcommand{\pmodel}{p_{\rm{model}}}
\newcommand{\Pmodel}{P_{\rm{model}}}
\newcommand{\ptildemodel}{\tilde{p}_{\rm{model}}}
\newcommand{\pencode}{p_{\rm{encoder}}}
\newcommand{\pdecode}{p_{\rm{decoder}}}
\newcommand{\precons}{p_{\rm{reconstruct}}}

\newcommand{\laplace}{\mathrm{Laplace}} 

\newcommand{\E}{\mathbb{E}}
\newcommand{\Ls}{\mathcal{L}}
\newcommand{\R}{\mathbb{R}}
\newcommand{\emp}{\tilde{p}}
\newcommand{\lr}{\alpha}
\newcommand{\reg}{\lambda}
\newcommand{\rect}{\mathrm{rectifier}}
\newcommand{\softmax}{\mathrm{softmax}}
\newcommand{\sigmoid}{\sigma}
\newcommand{\softplus}{\zeta}
\newcommand{\KL}{D_{\mathrm{KL}}}
\newcommand{\Var}{\mathrm{Var}}
\newcommand{\standarderror}{\mathrm{SE}}
\newcommand{\Cov}{\mathrm{Cov}}
\newcommand{\normlzero}{L^0}
\newcommand{\normlone}{L^1}
\newcommand{\normltwo}{L^2}
\newcommand{\normlp}{L^p}
\newcommand{\normmax}{L^\infty}

\newcommand{\parents}{Pa} 

\let\ab\allowbreak

%% file: introduction.tex
{G}{raph-structured} data is ubiquitous in the real world, such as social networks~\cite{Fan2019GraphNN,zhao2020semi,xu2018enhancing}, molecular structures~\cite{Mansimov2019MolecularGP,Chereda2019UtilizingMN} and knowledge graphs~\cite{Sorokin2018ModelingSW,liu2022federated}. With the growing interest in learning from graphs, graph neural networks (GNNs) are receiving more and more attention over the years. Generally, GNNs adopt message-passing mechanisms, which recursively propagate and fuse messages from neighbor nodes on the graphs. Hence, the learned node representation captures both node attributes and neighborhood information, which facilitates various downstream tasks such as node classification~\cite{kipf2016semi,velivckovic2017graph,hamilton2017inductive,ren2017robust}, graph classification~\cite{xu2018powerful}, and link prediction~\cite{zhang2018link,lu2022graph,ren2018tracking}.

Despite the success of GNNs for various domains, as with other neural networks, GNNs lack interpretability. 
Understanding the inner working of GNNs can bring several benefits. First, it enhances practitioners' trust in the GNN model by enriching their understanding of the model characteristics such as if the model is working as desired. Second, it increases the models' transparency to enable trusted applications in decision-critical fields sensitive to fairness, privacy and safety challenges~\cite{rao2021quantitative,ren2022mitigating,ren2021cross}. Thus, studying the explainability of GNNs is attracting increasing attention and many efforts have been taken~\cite{ying2019gnnexplainer,luo2020parameterized,zhao2022consistency}. 

Particularly, we focus on post-hoc instance-level explanations. Given a trained GNN and an input graph, 
this task seeks to discover the substructures that can explain the prediction behavior of the GNN model. Some solutions have been proposed in existing works~\cite{ying2019gnnexplainer,huang2020graphlime,vu2020pgm}. For example, in search of important substructures that predictions rely upon, GNNExplainer learns an importance matrix on node attributes and edges via perturbation~\cite{ying2019gnnexplainer}. The identified minimal substructures that preserve original predictions are taken as the explanation. Extending this idea, PGExplainer trains a graph generator to utilize global information in explanation and enable faster inference in the inductive setting~\cite{luo2020parameterized}. SubgraphX constraints explanations as connected subgraphs and conduct Monte Carlo tree search based on Shapley value~\cite{yuan2021explainability}. These methods can be summarized in a label preserving framework, i.e., the candidate explanation is formed as a masked version of the original graph and is identified as the minimal discriminative substructure.

\begin{figure}[t!]
  \centering
  \subfigure[Run 1]{
		\includegraphics[width=0.35\textwidth]{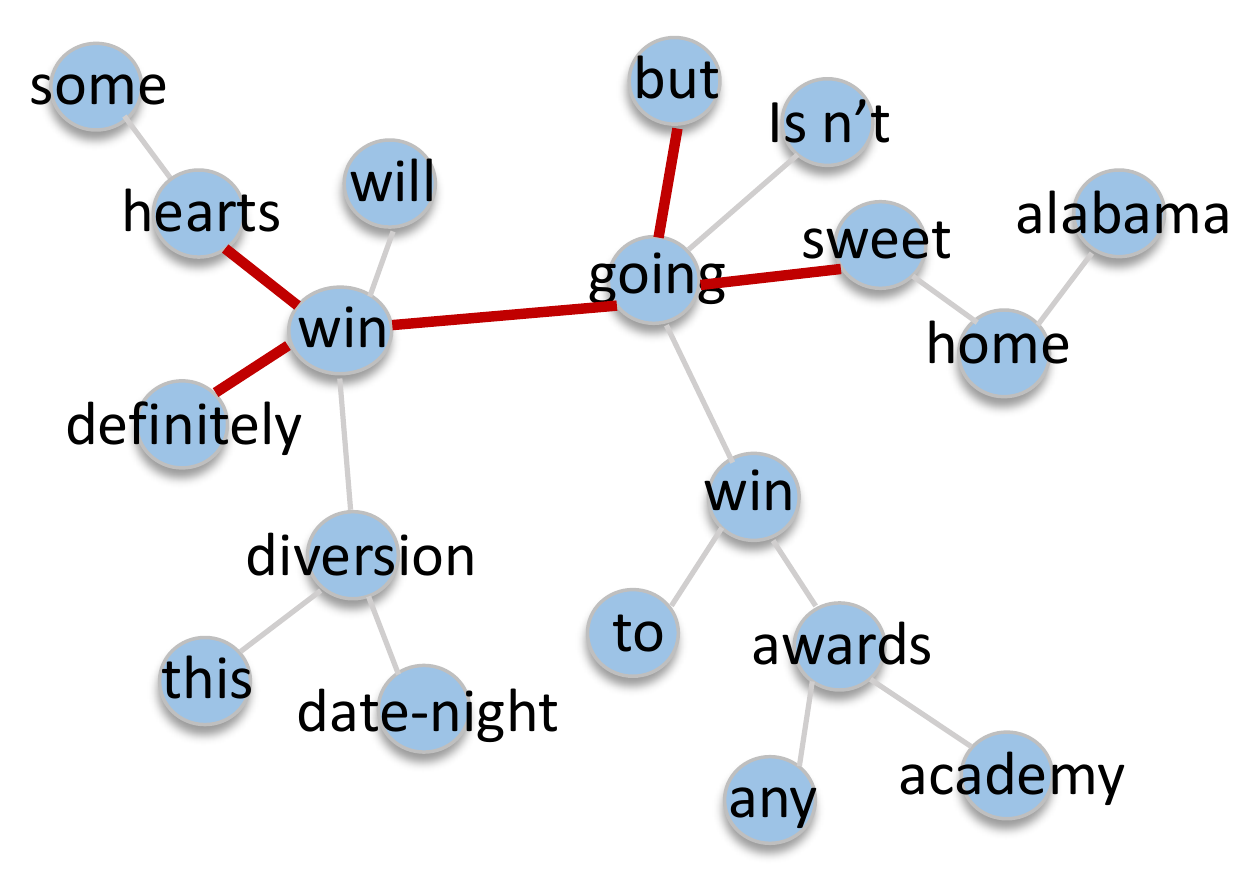}}
  \subfigure[Run 2]{
		\includegraphics[width=0.35\textwidth]{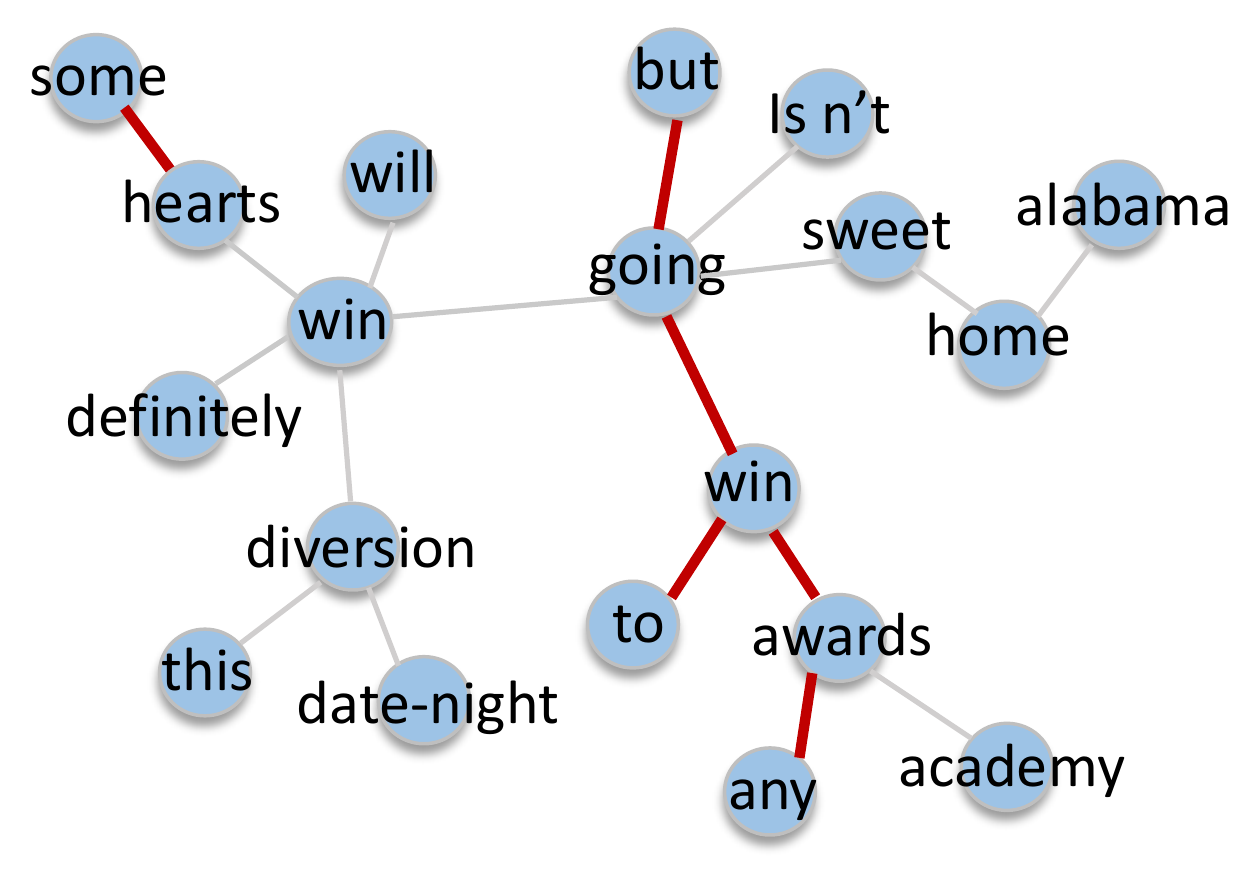}}
    \vskip -1em
    \caption{Explanation results achieved by a leading baseline GNNExplainer on the same input graph from Graph-SST5. Red edges formulate  explanation substructures.} \label{fig:case_study}
    \vskip -1.5em
\end{figure}

However, due to the complexity of topology and the combinatory number of candidate substructures, existing label preserving methods are insufficient for a faithful and consistent explanation of GNNs. They are unstable and are prone to give spurious correlations as explanations.
A failure case is shown in Figure~\ref{fig:case_study}, where a GNN is trained on Graph-SST5~\cite{yuan2020explainability} for sentiment classification. Each node represents a word and each edge denotes syntactic dependency between nodes. Each graph is labeled based on the sentiment of the sentence. In the figure, the sentence ``Sweet home alabama isn't going to win any academy awards, but this date-night diversion will definitely win some hearts'' is labeled \textit{positive}. In the first run, GNNExplainer~\cite{ying2019gnnexplainer} identifies the explanation as ``definitely win some hearts'', and in the second run, it identifies ``win academy awards'' to be the explanation instead. Different explanations obtained by GNNExplainer break the criteria of \textbf{consistency}, i.e., the explanation method should be deterministic and consistent with the same input for different runs~\cite{nauta2022anecdotal}. Consequently, explanations provided by existing methods may fail to faithfully reflect the decision mechanism of the to-be-explained GNN. 



Inspecting the inference process of target GNNs, we find that the inconsistency problem and spurious explanations can be understood from the causality perspective. Specifically, existing explanation methods may lead to spurious explanations either as a result of different causal factors or due to the confounding effect of distribution shifts (identified subgraphs may be out of distribution). These failure cases originate from a particular inductive bias that predicted labels are sufficiently indicative for extracting critical input components. This underlying assumption is rooted in optimization objectives adopted by existing works~\cite{ying2019gnnexplainer,luo2020parameterized,yuan2021explainability}. However, our analysis demonstrates that the label information is insufficient to filter out spurious explanations, leading to inconsistent and unfaithful explanations. 

Considering the inference of GNNs, both confounding effects and distinct causal relationships can be reflected in the internal representation space. With this observation, we propose a novel objective that encourages alignment of embeddings of raw graph and identified subgraph in internal embedding space to obtain more faithful and consistent GNN explanations. 
\tianxiang{Specifically, to evaluate the semantic similarity between two inputs and incorporate the alignment objective into explanation, we design and compare strategies with various design choices to measure similarity in the embedding space. These strategies enable the alignment between candidate explanations and original inputs and are flexible to be incorporated into various existing GNN explanation methods. Particularly, aside from directly using Euclidean distance, we further propose three distribution-aware strategies. The first one identifies a set of anchoring embeddings and utilizes relative distances against them. The second one assumes a Gaussian mixture model and captures the distribution using the probability of falling into each Gaussian center. The third one learns a deep neural network to estimate mutual information between two inputs, which takes a data-driven approach with little reliance upon prior domain knowledge.}
Further analysis shows that the proposed method is in fact optimizing a new explanation framework, which is more faithful in design. Our main contributions are:
\begin{itemize}[leftmargin=*]
    \item We point out the faithfulness and consistency issues in rationales identified by existing GNN explanation models. These issues arise due to the inductive bias in their label-preserving framework, which only uses predictions as the guiding information;
    \item 
    \tianxiang{We propose an effective and easy-to-apply countermeasure by aligning intermediate embeddings. We implement a set of variants with different alignment strategies, which is flexible to be incorporated to various GNN explanation models. }
    We further conduct a theoretical analysis to understand and validate the proposed framework. 
    \item Extensive experiments on real-world and synthetic datasets show that our framework benefits various GNN explanation models to achieve more faithful and consistent explanations. 
\end{itemize}

%% file: related_work.tex
\section{Related Work}
In this section, we review related works, including graph neural networks and interpretability of GNNs.


\subsection{Graph Neural Networks}
Graph neural networks (GNNs) are developing rapidly in recent years, with the increasing need for learning on relational data structures~\cite{Fan2019GraphNN,dai2022towards,zhao2021graphsmote,zhou2022graph,zhao2023skill}. Generally, 
existing GNNs can be categorized into two categories, i.e., spectral-based approaches~\cite{bruna2013spectral,liang2023abslearn,liang2022reasoning} based on graph signal processing theory, and spatial-based approaches~\cite{duvenaud2015convolutional,atwood2016diffusion,xiao2021learning} relying upon neighborhood aggregation. Despite their differences, most GNN variants can be summarized with the message-passing framework, which is composed of pattern extraction and interaction modeling within each layer~\cite{gilmer2017neural}. Specifically, GNNs model messages from node representations. These messages are then propagated with various message-passing mechanisms to refine node representations, which are then utilized for downstream tasks~\cite{hamilton2017inductive,zhang2018link,zhao2021graphsmote,zhao2022topoimb}. Explorations are made by disentangling the propagation process~\cite{wang2020disentangled,zhao2022exploring,xiao2022decoupled} or utilizing external prototypes~\cite{lin2022prototypical,xu2022hp}. Research has also been conducted on the expressive power~\cite{balcilar2021breaking,sato2020survey} and potential biases introduced by different kernels~\cite{nt2019revisiting,balcilar2021analyzing} for the design of more effective GNNs. Despite their success in network representation learning, GNNs are uninterpretable black box models. It is challenging to understand their behaviors even if the adopted message passing mechanism and parameters are given. Besides, unlike traditional deep neural networks where instances are identically and independently distributed, GNNs consider node features and graph topology jointly, making the interpretability problem more challenging to handle. 

\subsection{GNN Interpretation Methods}
Recently, some efforts have been taken to interpret GNN models and provide explanations for their predictions~\cite{wang2020causal}. Based on the granularity, existing methods can be generally grouped into two categories: (1) instance-level explanation~\cite{ying2019gnnexplainer}, which provides explanations on the prediction for each instance by identifying important substructures; and (2) model-level explanation~\cite{baldassarre2019explainability,zhang2021protgnn}, which aims to understand global decision rules captured by the target GNN. From the methodology perspective, existing methods can be categorized as (1) self-explainable GNNs~\cite{zhang2021protgnn,dai2021towards}, where the GNN can simultaneously give prediction and explanations on the prediction; and (2) post-hoc explanations~\cite{ying2019gnnexplainer,luo2020parameterized,yuan2021explainability}, which adopt another model or strategy to provide explanations of a target GNN. As post-hoc explanations are model-agnostic, i.e., can be applied for various GNNs, in this work, we focus on post-hoc instance-level explanations~\cite{ying2019gnnexplainer}, i.e., given a trained GNN model, identifying instance-wise critical substructures for each input to explain the prediction. A comprehensive survey can be found in ~\cite{dai2022comprehensive}.

A variety of strategies for post-hoc instance-level explanations have been explored in the literature, including utilizing signals from gradients based~\cite{pope2019explainability,baldassarre2019explainability}, perturbed predictions based~\cite{ying2019gnnexplainer,luo2020parameterized,yuan2021explainability,shan2021reinforcement}, and decomposition based~ \cite{baldassarre2019explainability,schnake2020higher}. Among these methods, perturbed prediction-based methods are the most popular. The basic idea is to learn a perturbation mask that filters out non-important connections and identifies dominating substructures preserving the original predictions~\cite{yuan2020explainability}. The identified important substructure is used as an explanation for the prediction. For example, GNNExplainer~\cite{ying2019gnnexplainer} employs two soft mask matrices on node attributes and graph structure, respectively, which are learned end-to-end under the maximizing mutual information (MMI) framework. PGExplainer~\cite{luo2020parameterized} extends it by incorporating a graph generator to utilize global information. It can be applied in the inductive setting and prevent the onerous task of re-learning from scratch for each to-be-explained instance. SubgraphX~\cite{yuan2021explainability} expects explanations to be in the form of sub-graphs instead of bag-of-edges and employs Monte Carlo Tree Search (MCTS) to find connected subgraphs that preserve predictions measured by the Shapley value. To promote faithfulness in identified explanations, some works introduced terminologies from the causality analysis domain, via estimating the individual causal effect of each edge~\cite{lin2021generative} or designing interventions to prevent the discovery of spurious correlations~\cite{wu2022discovering}. Ref.~\cite{yu2020graph} connects the idea of identifying minimally-predictive parts in explanation with the principle of information bottleneck~\cite{tishby2015deep} and designs an end-to-end optimization framework for GNN explanation.

Despite the aforementioned progress in interpreting GNNs, most of these methods discover critical substructures merely upon the change of outputs given perturbed inputs. Due to this underlying inductive bias, existing label-preserving methods are heavily affected by spurious correlations caused by confounding factors in the environment. On the other hand, by aligning intermediate embeddings in GNNs, our method alleviates the effects of spurious correlations on interpreting GNNs, leading to faithful and consistent explanations. 

\subsection{Graph Contrastive Learning}

\revision{
In recent years, contrastive learning (CL) has garnered significant attention as it mitigates the need for manual annotations via unsupervised pretext tasks~\cite{LKhc2020ContrastiveRL,Khosla2020SupervisedCL,chen2020simple}. In a typical CL framework, the model is trained in a pairwise manner, promoting attraction between positive sample pairs and repulsion between negative sample pairs~\cite{Graf2021DissectingSC,Wang2020UnderstandingCR}.

Contrastive Learning (CL) techniques have recently been extended to the graph domain, constructing multiple graph views and maximizing mutual information (MI) among semantically similar instances~\cite{chen2020simple,feng2022adversarial,xia2022hypergraph,zhu2021empirical,wang2022explanation,liang2023knowledge}. Existing methods primarily vary on graph augmentation techniques and contrastive pretext tasks. Graph augmentations commonly involve node-level attribute masking or perturbation~\cite{Hu2019StrategiesFP,Opolka2019SpatioTemporalDG,Velickovic2018DeepGI}, edge dropping or rewiring~\cite{Zhu2020SelfsupervisedTO,You2020GraphCL}, and graph diffusions\cite{Hassani2020ContrastiveMR,Jin2021MultiScaleCS}. Contrastive pretext tasks can be categorized into two branches, \textit{same-scale} contrasts between embeddings of node (graph) pairs~\cite{Zhu2020DeepGC,Hassani2020ContrastiveMR,Zeng2022ImGCLRG} and \textit{cross-scale} contrasts between global graph embeddings and local node representations~\cite{Velickovic2018DeepGI,Opolka2019SpatioTemporalDG}. Recent works have also explored dynamic contrastive objectives~\cite{You2021GraphCL,Yin2021AutoGCLAG} via learning the augmentation strategies adaptively. }

%% file: problem_setting.tex
\section{Preliminary}

\subsection{Problem Definition}
We use $\mathcal{G} = \{\mathcal{V},\mathcal{E}; \mathbf{F}, \mathbf{A} \}$ to denote a graph, where $\mathcal{V}=\{v_1,\dots,v_{n}\}$ is a set of $n$ nodes and $\mathcal{E} \in \mathcal{V} \times \mathcal{V}$ is the set of edges. Nodes are accompanied by an attribute matrix $\mathbf{F} \in \mathbb{R}^{n \times d}$, and $\mathbf{F}[i,:] \in \mathbb{R}^{1 \times d}$ is the $d$-dimensional node attributes of node $v_i$. $\mathcal{E}$ is described by an adjacency matrix $\mathbf{A} \in \mathbb{R}^{n \times n}$. ${A}_{ij}=1$ if there is an edge between node $v_i$ and $v_j$; otherwise, ${A}_{ij}=0$. For \textit{graph classification}, each graph $\mathcal{G}_i$ has a label $Y_i \in \mathcal{C}$, and a GNN model $f$ is trained to map $\mathcal{G}$ to its class, i.e., $f: \{\mathbf{F}, \mathbf{A}\} \mapsto \{1, 2, \dots, C \}$. Similarly, for \textit{node classification}, each graph $\mathcal{G}_i$  denotes a $K$-hop subgraph centered at node $v_i$ and a GNN model $f$ is trained to predict the label of $v_i$ based on node representation of $v_i$ learned from $\mathcal{G}_i$. The purpose of explanation is to find a subgraph $\mathcal{G}'$, marked with binary importance mask $\mathbf{M}_{A} \in [0,1]^{n \times n}$ on adjacency matrix and $\mathbf{M}_{F} \in[0,1]^{n \times d}$ on node attributes, respectively, e.g., $\mathcal{G}'=\{\mathbf{A}\odot \mathbf{M}_A; \mathbf{F}\odot \mathbf{M}_F\}$, where $\odot$ denotes elementwise multiplication. 
These two masks highlight components of $\mathcal{G}$ that are important for $f$ to predict its label. With the notations, the \textit{post-hoc instance-level} GNN explanation task is:

\vspace{0.5em}
\noindent{}\textit{Given a trained GNN model $f$, for an arbitrary input graph $\mathcal{G}= \{\mathcal{V}, \mathcal{E};  \mathbf{F}, \mathbf{A}\}$, find a subgraph $\mathcal{G}'$ that can explain the prediction of $f$ on $\mathcal{G}$. The obtained explanation $\mathcal{G}'$ is depicted by importance mask $\mathbf{M}_{F}$ on node attributes and importance mask $\mathbf{M}_{A}$ on adjacency matrix.
}

\subsection{ MMI-based Explanation Framework}
Many approaches have been designed for post-hoc instance-level  GNN explanation. Due to the discreteness of edge existence and non-grid graph structures, most works apply a perturbation-based strategy to search for explanations. Generally, they can be summarized as Maximization of Mutual Information (MMI) between predicted label $\hat{Y}$ and explanation $\mathcal{G}'$, i.e.,
\begin{equation}\label{eq:framework}
    \begin{aligned}
    \min_{\mathcal{G}'} & - I(\hat{Y}, \mathcal{G}'), \\
    \quad \text{s.t.} \quad \mathcal{G}' \sim & \mathcal{P}(\mathcal{G}, \mathbf{M}_{A},  \mathbf{M}_{F}),  \quad \mathcal{R}(\mathbf{M}_{F},\mathbf{M}_{A}) \leq c 
    \end{aligned}
\end{equation}
where $I()$ represents mutual information and $\mathcal{P}$ denotes the perturbations on original input with importance masks $\{\mathbf{M}_{F},\mathbf{M}_{A}\}$. For example, let $\{\hat{\mathbf{A}},\hat{\mathbf{F}}\}$ represent the perturbed $\{\mathbf{A}, \mathbf{F}\}$. Then, $\hat{\mathbf{A}}= \mathbf{A} \odot \mathbf{M}_{A}$ and $\hat{\mathbf{F}} = \mathbf{Z} + (\mathbf{F}-\mathbf{Z}) \odot \mathbf{M}_{F}$ in GNNExplainer~\cite{ying2019gnnexplainer}, where $\mathbf{Z}$ is sampled from marginal distribution of node attributes $\mathbf{F}$. $\mathcal{R}$ denotes regularization terms on the explanation, imposing prior knowledge into the searching process, like constraints on budgets or connectivity distributions~\cite{luo2020parameterized}. Mutual information $I(\hat{Y}, \mathcal{G}')$ quantifies consistency between original predictions $\hat{Y}=f(\mathcal{G})$ and prediction of candidate explanation $f(\mathcal{G}')$, which promotes the positiveness of found explanation $\mathcal{G}'$. 
Since mutual information measures the predictive power of $\mathcal{G}'$ on $Y$, this framework essentially tries to find a subgraph that can best predict the original output $\hat{Y}$. During training, a relaxed version~\cite{ying2019gnnexplainer} is often adopted as: 
\begin{equation}\label{eq:surrogate}
    \begin{aligned}
    \min_{\mathcal{G}'} ~& {H}_{C}\big(\hat{Y}, P(\hat{Y}' \mid \mathcal{G}') \big),\\
     \quad \text{s.t.} \quad \mathcal{G}' \sim & \mathcal{P}(\mathcal{G}, \mathbf{M}_{A},  \mathbf{M}_{F}) , \quad \mathcal{R}(\mathbf{M}_{F},\mathbf{M}_{A}) \leq c 
    \end{aligned}
\end{equation}
where $H_C$ denotes cross-entropy. With this same objective, existing methods mainly differ from each other in optimization and searching strategies.


Different aspects regarding the quality of explanations can be evaluated~\cite{nauta2022anecdotal}. Among them, two most important criteria are \textbf{faithfulness} and \textbf{consistency}. Faithfulness measures the descriptive accuracy of explanations, indicating how truthful they are compared to behaviors of the target model. Consistency considers explanation invariance, which checks that identical input should have identical explanations. However, as shown in Figure~\ref{fig:case_study}, the existing MMI-based framework is sub-optimal in terms of these criteria. The cause of this problem is rooted in its learning objective, which uses prediction alone as guidance in search of explanations. Due to the complex graph structure, the prediction alone as a guide could result in spurious explanations. A detailed analysis will be provided in the next section.

\section{Analyze Spurious Explanations}~\label{sec:analysis}
\begin{figure}[t]
    \centering

  \subfigure[SCM]{    \label{fig:SEM}
		\includegraphics[width=0.29\textwidth]{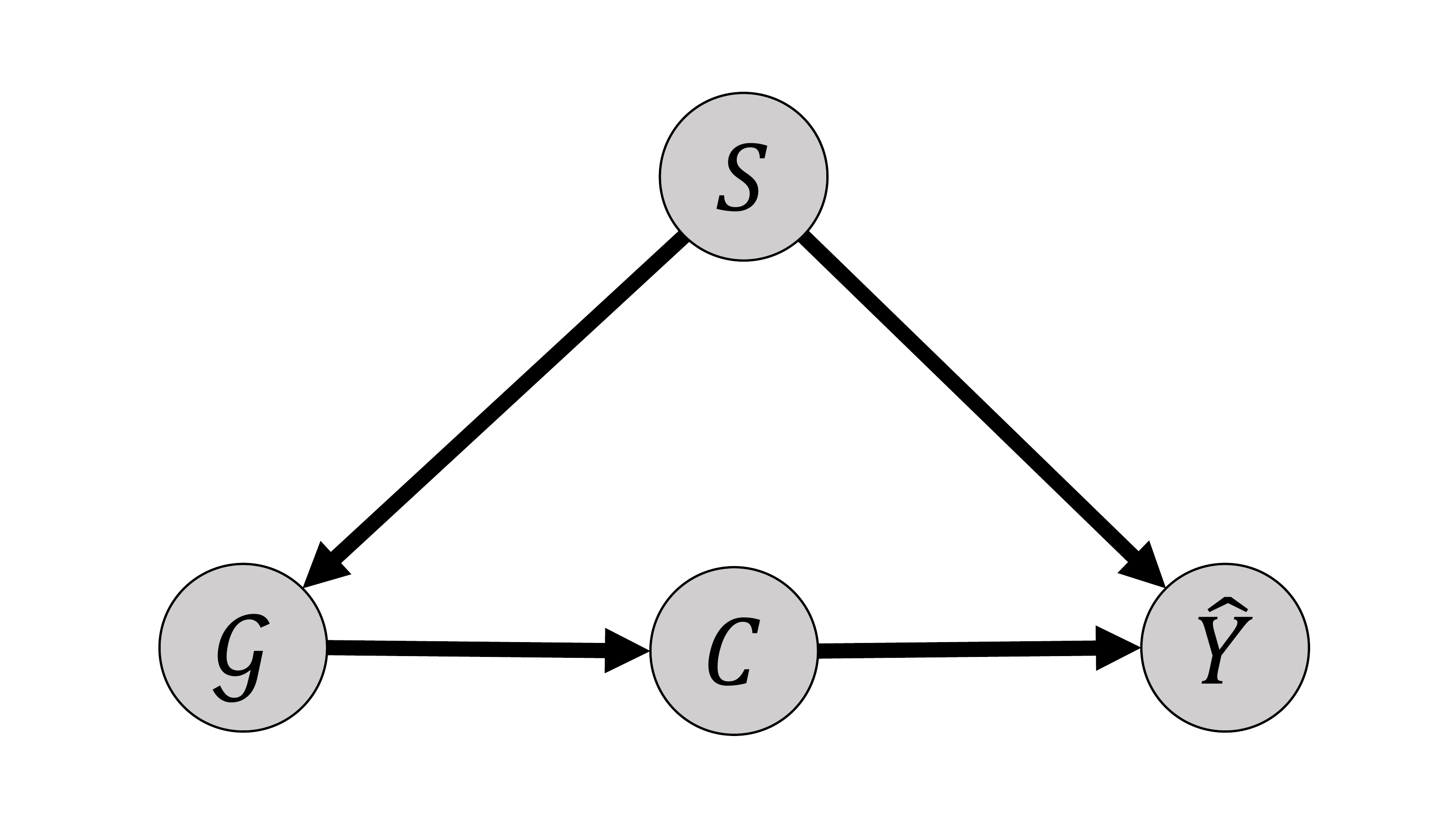}}
  \subfigure[Alignment]{\label{fig:alignment}
		\includegraphics[width=0.26\textwidth]{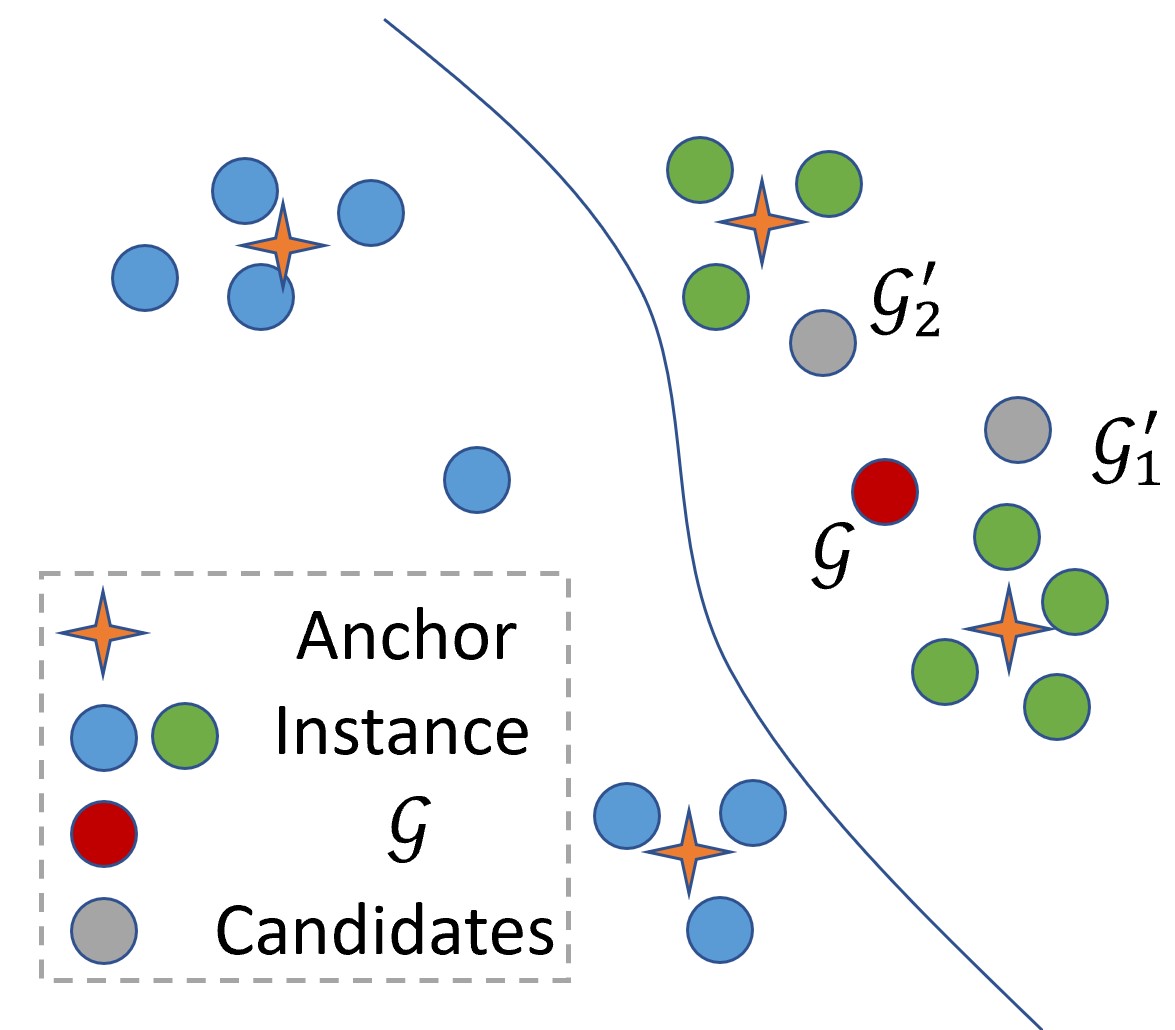}}
    \vskip -1em
    \caption{(a) Prediction rules of $f$, in the form of SCM. (b) An example of anchor-based embedding alignment.}

    \vskip -1.5em
\end{figure}

With ``spurious explanations'', we refer to those explanations lie outside the genuine rationale of prediction on $\mathcal{G}$, making the usage of $\mathcal{G}'$ as explanations anecdotal. 
As examples in Figure~\ref{fig:case_study}, despite rapid developments in explaining GNNs, the problem w.r.t faithfulness and consistency of detected explanations remains. To get a deeper understanding of reasons behind this problem, we will examine the behavior of target GNN model from the causality perspective. Figure ~\ref{fig:SEM} shows the Structural Equation Model (SEM), where variable $C$ denotes discriminative causal factors and variable $S$ represents confounding environment factors. Two paths between $\mathcal{G}$ and the predicted label $\hat{Y}$ can be found.
\begin{itemize}[leftmargin=*]
    \item $\mathcal{G} \rightarrow C \rightarrow \hat{Y}$: This path presents the inference of target GNN model, i.e., critical patterns $C$ that are informative and discriminative for the prediction $\hat{Y}$ would be extracted from input graph, upon which the target model is dependent. Causal variables are determined by both the input graph and learned knowledge by the target GNN model. 
    \item $\mathcal{G} \leftarrow S \rightarrow \hat{Y}$: We denote $S$ as the confounding factors, such as depicting the overall distribution of graphs. It is causally related to both the appearance of input graphs and the prediction of target GNN models. A masked version of $\mathcal{G}$ could create out-of-distribution (OOD) examples, resulting in spurious causality to prediction outputs. For example in the chemical domain, removing edges (bonds) or nodes (atoms) may obtain invalid molecular graphs that never appear during training. In the existence of distribution shifts, model predictions would be less reliable.
\end{itemize}

Figure~\ref{fig:SEM} provides us with a tool to analyze $f$'s behaviors. From the causal structures, we can observe that spurious explanations may arise as a result of failure in recovering the original causal rationale. $\mathcal{G}'$ learned from Equation~\ref{eq:framework} may preserve prediction $\hat{Y}$ due to confounding effect of distribution shift or different causal variables $C$ compared to original $\mathcal{G}$. Weakly-trained GNN $f(\cdot)$ that are unstable or non-robust towards noises would further amplify this problem as the prediction is unreliable. 

To further understand the issue, we build the correspondence from SEM in Figure~\ref{fig:SEM} to the inference process of GNN $f$. Specifically, we first decompose $f()$ as a feature extractor $f_{ext}()$ and a classifier $f_{cls}()$. Then, its inference can be summarized as two steps: (1) encoding step with $f_{ext}()$, which takes $\mathcal{G}$ as input and produce its embedding in the representation space $E_C$; (2) classification step with $f_{cls}()$,  which predicts output labels on input's embedding. Connecting these inference steps to SEM in Figure~\ref{fig:SEM}, we can find that: 
\begin{itemize}[leftmargin=*]
    \item The causal path $\mathcal{G} \rightarrow C \rightarrow \hat{Y} $ lies behind the inference process with representation space $E_C$ to encode critical variables $C$;
    \item The confounding effect of distribution shift $S$ works on the inference process via influencing distribution of graph embedding in $E_C$. When masked input $\mathcal{G}'$ is OOD, its embedding would fail to reflect its discriminative features and deviate from real distributions, hence deviating the classification step on it.
\end{itemize}
To summarize, we can observe that spurious explanations are usually obtained due to the following two reasons:
\begin{enumerate}[leftmargin=*]
    \item The obtained $\mathcal{G}'$ is OOD graph. During inference of target GNN model, the encoded representation of $\mathcal{G}'$ is distant from those seen in the training set, making the prediction unreliable;
    \item The encoded discriminative representation does not accord with that of the original graph. Different causal factors ($C$) are extracted between $\mathcal{G}'$ and $\mathcal{G}$, resulting in false explanations. 
\end{enumerate}

%% file: methodology.tex
\section{Methodology}
Based on the discussion above, in this section, we focus on improving the faithfulness and consistency of GNN explanations and correcting the inductive bias caused by simply relying on prediction outputs. 
We first provide an intuitive introduction to the proposed countermeasure, which takes the internal inference process into account. We then design four concrete algorithms to align $\mathcal{G}$ and $\mathcal{G}'$ in the latent space, to promote that they are seen and processed in the same manner. Finally, theoretical analysis is provided to justify our strategies.

\subsection{Alleviate Spurious Explanations}
Instance-level post-hoc explanation dedicates to finding discriminative substructures that the target model $f$ depends upon. The traditional objective in Equation~\ref{eq:surrogate} can identify minimal predictive parts of input, however, it is dangerous to directly take them as explanations. Due to diversity in graph topology and combinatory nature of sub-graphs, multiple distinct substructures could be identified leading to the same prediction, as discussed in Section~\ref{sec:analysis}.


For an explanation substructure $\mathcal{G}'$ to be faithful, it should follow the same rationale as the original graph $\mathcal{G}$ inside the internal inference of to-be-explained model $f$. To achieve this goal, the explanation $\mathcal{G}'$ should be aligned to $\mathcal{G}$ w.r.t the decision mechanism, reflected in Figure~\ref{fig:SEM}. However, it is non-trivial to extract and compare the critical causal variables $C$ and confounding variables $S$ due to the black box nature of the target GNN model to be explained.

Following the causal analysis in Section~\ref{sec:analysis}, we propose to take an alternative approach by looking into internal embeddings learned by $f$. Causal variables $C$ are encoded in representation space extracted by $f$, and out-of-distribution effects can also be reflected by analyzing embedding distributions. 
An assumption can be safely made: \textit{if two graphs are mapped to embeddings near each other by a GNN layer, then these graphs are seen as similar by it and would be processed similarly by following layers}. 
With this assumption, a proxy task can be designed by aligning internal graph embeddings between $\mathcal{G}'$ and $\mathcal{G}$. This new task can be incorporated into Framework~\ref{eq:framework} as an auxiliary optimization objective.

Let $\mathbf{h}_v^l$ be the representation of node $v$ at the $l$-th GNN layer with $\mathbf{h}_v^0 = \mathbf{F}[v,:]$. Generally, the inference process inside GNN layers can be summarized as a message-passing framework:
    \begin{equation}\label{eq:messagepass}
        \begin{aligned}
        \mathbf{m}_{v}^{l+1} &=\sum_{u \in \mathcal{N}(v)} \text{Message}_{l}\large( \mathbf{h}_v^l, \mathbf{h}_u^l, A_{v,u} \large), \\
        \mathbf{h}_v^{l+1} &= \text{Update}_{l}\large(\mathbf{h}_v^{l}, \mathbf{m}_{v}^{l+1} \large),
        \end{aligned}
    \end{equation}
where $\text{Message}_{l}$ and $\text{Update}_{l}$ are the message function and update function at $l$-th layer, respectively. $\mathcal{N}(v)$ is the set of node $v$'s neighbors. Without loss of generality, the graph pooling layer can also be presented as:
    \begin{equation}\label{eq:gpool}
        \mathbf{h}_{v'}^{l+1} = \sum_{v\in \mathcal{V}} {P}_{v, v'} \cdot \mathbf{h}_{v}^{l}.
    \end{equation}
where ${P}_{v,v'}$ denotes mapping weight from node $v$ in layer $l$ to node $v'$ in layer $l+1$ inside the myriad of GNN for graph classification. We propose to align embedding $\mathbf{h}_v^{l+1}$ at each layer, which contains both node and local neighborhood information.

\subsection{Distribution-Aware Alignment}\label{sec:implement}

Achieving alignment in the embedding space is not straightforward. It has several distinct difficulties. (1) It is difficult to evaluate the distance between $\mathcal{G}$ and $\mathcal{G}'$ in this embedding space. Different dimensions could encode different features and carry different importance. Furthermore, $\mathcal{G}'$ is a substructure of the original $\mathcal{G}$, and a shift on unimportant dimensions would naturally exist. (2) Due to the complexity of graph/node distributions, it is non-trivial to design a measurement of alignments that is both computation-friendly and can correlate well to distance on the distribution manifold. 

To address these challenges, we design a strategy to identify explanatory substructures and preserve their alignment with original graphs in a distribution-aware manner. The basic idea is to utilize other graphs to obtain a global view of the distribution density of embeddings, providing a better measurement of alignment. Concretely, we obtain representative node/graph embeddings as anchors and use distances to these anchors as the distribution-wise representation of graphs. Alignment is conducted on obtained representation of graph pairs. Next, we go into details of this strategy.
    \begin{itemize}[leftmargin=0.2in]
        \item  First, using graphs $\{\mathcal{G}_i\}_{i=1}^{m}$ from the same dataset, a set of node embeddings can be obtained as $\{\{\mathbf{h}^{l}_{v,i}\}_{v \in \mathcal{V}'_i}\}_{i=1}^{m}$ for each layer $l$, where $\mathbf{h}_{v,i}$ denotes embedding of node $v$ in graph $\mathcal{G}_i$. For node-level tasks, we set $\mathcal{V}'_i$ to contain only the center node of graph $\mathcal{G}_i$. For graph-level tasks, $\mathcal{V}'_i$ contains nodes set after graph pooling layer, and we process them following $\{ \sum_{v \in \mathcal{V}'_i} \mathbf{h}^{l+1}_{v,i}/|\mathcal{V}'_i| \}_{i=1}^{m}$ to get global graph representation.
        \item Then, a clustering algorithm is applied to the obtained embedding set to get $K$ groups. Clustering centers of these groups are set to be anchors, annotated as $\{\mathbf{h}^{l+1,k}\}_{k=1}^{K}$. In experiments, we select DBSCAN~\cite{ester1996density} as the clustering algorithm, and tune its hyper-parameters to get around $20$ groups.
        \item At $l$-th layer, $\mathbf{h}_v^{l+1}$ is represented in terms of relative distances to those $K$ anchors, as $\mathbf{s}_v^{l} \in \mathbb{R}^{1 \times K}$ with the $k$-th element calculated as $\mathbf{s}_{v}^{l+1,k} = \|\mathbf{h}^{l+1}_{v}-\mathbf{h}^{l+1,k}_{v}\|_2$.
    \end{itemize}
Alignment between $\mathcal{G}'$ and $\mathcal{G}$ can be achieved by comparing their representations at each layer. The alignment loss is computed as:
    \begin{equation}\label{eq:old_align}
       \mathcal{L}_{align}\large( f(\mathcal{G}), f(\mathcal{G}') \large) = \sum_l  \sum_{v\in \mathcal{V}'} \| \mathbf{s}_{v}^l - {\mathbf{s}}_{v}^{'l} \|_2^2.
    \end{equation}
This metric provides a lightweight strategy for evaluating alignments in the embedding distribution manifold, by comparing relative positions w.r.t representative clustering centers. This strategy can naturally encode the varying importance of each dimension. 
Fig.~\ref{fig:alignment} gives an example, where $\mathcal{G}$ is the graph to be explained and the red stars are anchors. $\mathcal{G}'_1$ and $\mathcal{G}'_2$ are both similar to $\mathcal{G}$ w.r.t absolute distances; while it is easy to see $\mathcal{G}'_1$ is more similar to $\mathcal{G}$ w.r.t to the anchors. In other words, the anchors can better measure the alignment to filter out spurious explanations.

This alignment loss is used as an auxiliary task incorporated into MMI-based framework in Equation~\ref{eq:surrogate} to get faithful explanation as:
\begin{equation}
    \begin{aligned}
    \min_{\mathcal{G}'}  &H_C\big(\hat{Y}, P(\hat{Y}' \mid \mathcal{G}') \big) + \lambda \cdot \mathcal{L}_{Align}, \\
     \quad \text{s.t.} \quad \mathcal{G}' \sim & \mathcal{P}(\mathcal{G}, \mathbf{M}_{A},  \mathbf{M}_{F}), \quad \mathcal{R}(\mathbf{M}_{F},\mathbf{M}_{A}) \leq c 
    \end{aligned}\label{eq:target}
\end{equation}
where $\lambda$ controls the balance between prediction preservation and embedding alignment.  $\mathcal{L}_{Align}$ is flexible to be incorporated into various existing explanation methods.

\subsection{Direct Alignment}
As a simpler and more direct implementation, we also design a variant based on absolute distance. For layers without graph pooling, the objective can be written as $\sum_l\sum_{v\in \mathcal{V}} \|\mathbf{h}_{v}^l - {\mathbf{h}}_{v}^{'l} \|_2^2 $. For layers with graph pooling, as the structure could be different, we conduct alignment on global representation $ \sum_{v \in \mathcal{V}'} \mathbf{h}^{l+1}_{v}/|\mathcal{V}'| $, where $\mathcal{V}'$ denotes node set after pooling. 

\section{Extended Methodology}
In this section, we further examine more design choices for the strategy of alignment to obtain faithful and consistent explanations. Instead of using heuristic approaches, we explore two new directions: (1) statistically sound distance measurements based on the Gaussian mixture model, (2) fully utilizing the power of deep neural networks to capture distributions in the latent embedding space. Details of these two alignment strategies will be introduced below. 

\subsection{Gaussian-Mixture-based Alignment}
In this strategy, we model the latent embeddings of nodes (or graphs) using a mixture of Gaussian distributions, with representative node/graph embeddings as anchors (Gaussian centers). The produced embedding of each input can be compared with those prototypical anchors, and the semantic information of inputs taken by the target model would be encoded by relative distances from them. 

Concretely, we first obtain prototypical representations, annotated as $\{\mathbf{h}^{l,k}\}_{k=1}^{K}$, by running the clustering algorithm on collected embeddings $\{\{\mathbf{h}^{l}_{v,i}\}_{v \in \mathcal{V}'_i}\}_{i=1}^{m}$ from graphs $\{\mathcal{G}_i\}_{i=1}^{m}$, in the same strategy as introduced in Sec.~\ref{sec:implement}. Clutering algorithm DBSCAN~\cite{ester1996density} is adopted and we tune its hyper-parameters to get around $20$ groups.

Next, the probability of encoded representation ${\mathbf{h}}^{l}$ falling into each prototypical Gaussian centers $\{\mathbf{h}^{l,k}\}_{k=1}^K$ can be computed as:
\begin{equation}\label{eq:GaussP}
    p_v^{l,k} = \frac{\exp(-\|{\mathbf{h}}_v^l-\mathbf{h}^{l,k}\|_2^2/2\sigma^2)}{\sum_{j=1}^K \exp(-\|{\mathbf{h}}_v^l-\mathbf{h}^{l,k}\|_2^2/2\sigma^2)}
\end{equation}
This distribution probability can serve as a natural tool for depicting the semantics of the input graph learned by the GNN model. Consequently, the distance between ${\mathbf{h}}_v^{'l}$ and $\mathbf{h}_v^l$ can be directly measured as the KL-divergence w.r.t this distribution probability:
\begin{equation}\label{eq:GaussD}
    d( {\mathbf{p}}_v^{'l}, \mathbf{p}_v^{l}) = \sum_{k \in [1, \dots, K]} {p}_v^{'l,k} \cdot \log(\frac{{p}_v^{'l,k}}{{p}_v^{l,k}}),
\end{equation}
where $\mathbf{p}_v^{'l} \in \mathbb{R}^K$ denotes the distribution probability of candidate explanation embedding, $\mathbf{h}_v^{'l}$. Using this strategy, the alignment loss between original graph and the candidate explanation is computed as:
\begin{equation}
    \mathcal{L}_{align}\big( f(\mathcal{G}), f(\mathcal{G}') \big) = \sum_l  \sum_{v\in \mathcal{V}'} d({\mathbf{p}}_v^{'l}, {\mathbf{p}_v^l}),
\end{equation}
which can be incorporated into the revised explanation framework proposed in Eq.~\ref{eq:target}. 

\textbf{Comparison} Compared with the alignment loss based on relative distances against anchors in Eq.~\ref{eq:old_align}, this new objective offers a better strategy in taking distribution into consideration. Specifically, we can show the following two advantages: 
\begin{itemize}[leftmargin=*]
    \item In obtaining the distribution-aware representation of each instance, this variant uses a Gaussian distance kernel (Eq.~\ref{eq:GaussP}) while the other one uses Euclidean distance in Sec.~\ref{sec:implement}, which may amplify the influence of distant anchors. We can prove this by examining the gradient of changes in representation w.r.t GNN embeddings $\mathbf{h}_v$. In the $l$-th layer at dimension $k$, the gradient of the previous variant can be computed as: 
    \begin{equation}
        \frac{\partial s_v^{l,k}}{\partial\mathbf{h}_v^{l}} = 2\cdot (\mathbf{h}_v^{l} - \mathbf{h}_v^{l,k})
    \end{equation}
    On the other hand, the gradient of this variant is:
    \begin{equation}
    \begin{aligned}
        \frac{\partial p_v^{l,k}}{\partial\mathbf{h}_v^{l}} & \approx  -\frac{\exp(-\|{\mathbf{h}}_v^l-\mathbf{h}^{l,k}\|_2^2/2\sigma^2)\cdot (\mathbf{h}_v^{l} - \mathbf{h}_v^{l,k})}{\sigma^2\cdot\sum_{j=1}^K \exp(-\|{\mathbf{h}}_v^l-\mathbf{h}^{l,k}\|_2^2/2\sigma^2)} 
    \end{aligned} 
    \end{equation}
    It is easy to see that for the previous variant, the magnitude of gradient would grow linearly w.r.t distances towards corresponding anchors. For this variant, on the other hand, the term $\frac{\exp(-\|{\mathbf{h}}_v^l-\mathbf{h}^{l,k}\|_2^2/2\sigma^2)}{\sigma^2\cdot\sum_{j=1}^K \exp(-\|{\mathbf{h}}_v^l-\mathbf{h}^{l,k}\|_2^2/2\sigma^2)}$ would down-weight the importance of those distant anchors while up-weight the importance of similar anchors, which is more desired in obtaining distribution-aware representations.
    \item In computing the distance between representations of two inputs, this variant adopts the KL divergence as in Eq.~\ref{eq:GaussD}, which would be scale-agnostic compared to the other one directly using Euclidean distance as in Eq.~\ref{eq:old_align}. Again, we can show the gradient of alignment loss towards obtained embeddings that encode distribution information. It can be observed that for the previous variant:
    \begin{equation}
        \frac{\partial d(\mathbf{s}_v^{l}, \mathbf{s}_v^{'l})}{\partial s_v^{'l,k}} = 2\cdot (s_v^{l,k}-s_v^{'l,k})
    \end{equation}
    For this variant based on Gaussian mixture model, the gradient can be computed as:
    \begin{equation}
        \frac{\partial d(\mathbf{p}_v^{l}, \mathbf{p}_v^{'l})}{\partial p_v^{'l,k}} =  1 + \log(\frac{{p}_v^{'l,k}}{{p}_v^{l,k}})
    \end{equation}
It can be observed that the previous strategy focuses on representation dimensions with a large absolute difference, while would be sensitive towards the scale of each dimension. On the other hand, this strategy uses the summation between the logarithm of relative difference with a constant, which is scale-agnostic towards each dimension.

\end{itemize}

\subsection{MI-based Alignment}
In this strategy, we further consider the utilization of deep models to capture the distribution and estimate the semantic similarity of two inputs, and incorporate it into the alignment loss for discovering faithful and consistent explanations. Specifically, we train a deep model to estimate the mutual information (MI) between two input graphs, and use its prediction as a measurement of alignment between original graph and its candidate explanation. This strategy circumvents the reliance on heuristic strategies and is purely data-driven, which can be learned in an end-to-end manner. 

To learn the mutual information estimator, we adopt a neural network and train it to be a Jensen-Shannon MI estimator~\cite{hjelm2018learning}. Concretely, we train this JSD-based estimator on top of intermediate embeddings with the learning objective as follows, which offers better stability in optimization: 
\begin{equation}
    \begin{aligned}\label{eq:Lmi}
    \min_{g_{mi}} \mathcal{L}_{mi} =& \E_{\mathcal{G}\in \{\mathcal{G}_i\}_{i=1}^m}\E_{v \in \mathcal{G}}\E_l[\E_{\mathbf{h}_v^{l,+}} sp(-T^l(\mathbf{h}_v^l, \mathbf{h}_v^{l,+})) \\
    &+ \E_{\mathbf{h}_v^{l,-}} sp(T^l(\mathbf{h}_v^l, \mathbf{h}_v^{l,-}))],
    \end{aligned}
\end{equation}
where $\E$ denotes expectation. In this equation, $T^l(\cdot)$ is a compatibility estimation function in the $l$-th layer, and we denote $\{T^l(\cdot)\}_l$ as the MI estimator $g_{mi}$. Activation function $sp(\cdot)$ is the \textit{softplus} function, and $\mathbf{h}_v^+$ represents the embedding of augmented node $v$ that is a \textbf{positive pair} of $v$ in the original graph. On the contrary, $\mathbf{h}_v^-$ denotes the embedding of augmented node $v$ that is a \textbf{negative pair} of original input. A positive pair is obtained through randomly dropping intermediate neurons, corresponding to masking out a ratio of original input, and a negative pair is obtained as embeddings of different nodes. This objective can guide $g_{mi}$ to capture the correlation or similarity between two input graphs encoded by the target model. \revision{Note that to further improve the learning of MI estimator, more advanced graph augmentation techniques can be incorporated for obtaining positive/negative pairs, following recent progresses in graph contrastive learning~\cite{feng2022adversarial,xia2022hypergraph}. In this work, we stick to the strategy adopted in ~\cite{hjelm2018learning} for its popularity, and leave that for future explorations.} With this MI estimator learned, alignment loss between $\mathcal{G}$ and $\mathcal{G}^{'}$ can be readily computed:
\begin{equation} \small
    \mathcal{L}_{align}\big( f(\mathcal{G}), f(\mathcal{G}') \big) = \sum_l  \sum_{v\in \mathcal{V}'} sp(-T^l({\mathbf{h}}_v^{'l}, \mathbf{h}_v^{l})),
\end{equation}
which can be incorporated into the revised explanation framework proposed in Eq.~\ref{eq:target}. 

In this strategy, we design a data-driven approach by capturing the mutual information between two inputs, which circumvents the potential biases of using human-crafted heuristics.

\section{Theoretical Analysis}
With those alignment strategies and our new explanation framework introduced, next, we want to look deeper and provide theoretical justifications for the proposed new loss function in Eq.~\ref{eq:target}. In this section, we first propose a new explanation objective to prevent spurious explanations based Sec.~\ref{sec:analysis}. Then, we theoretically show that it squares with our loss function with mild relaxations.

\subsection{New Explanation Objective}

\begin{figure}[t!]
  \centering
  \subfigure[Previous Objective]{\label{fig:venn_a}
		\includegraphics[width=0.33\textwidth]{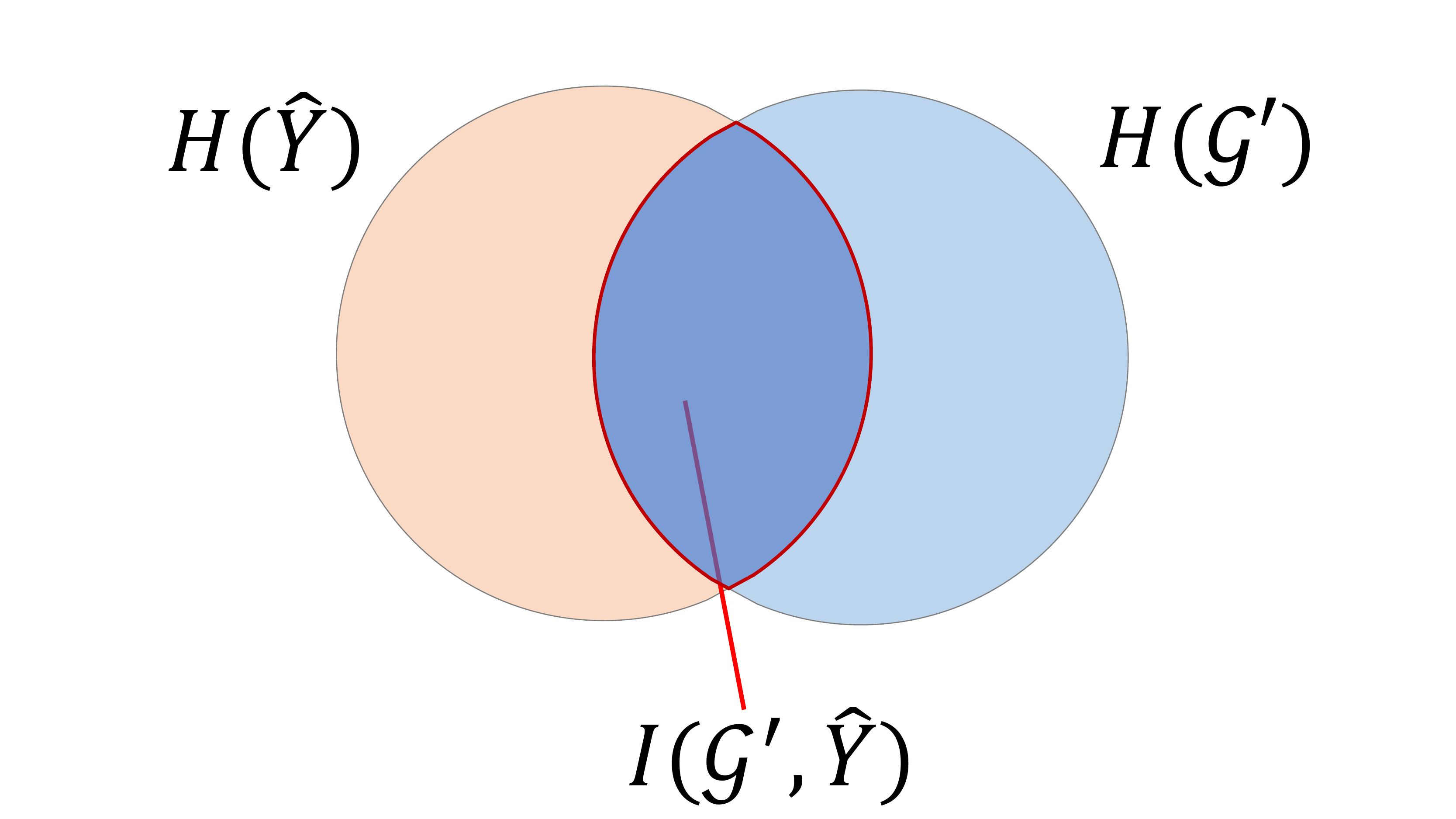}}
  \subfigure[Our Proposed New Objective]{\label{fig:venn_b}
		\includegraphics[width=0.35\textwidth]{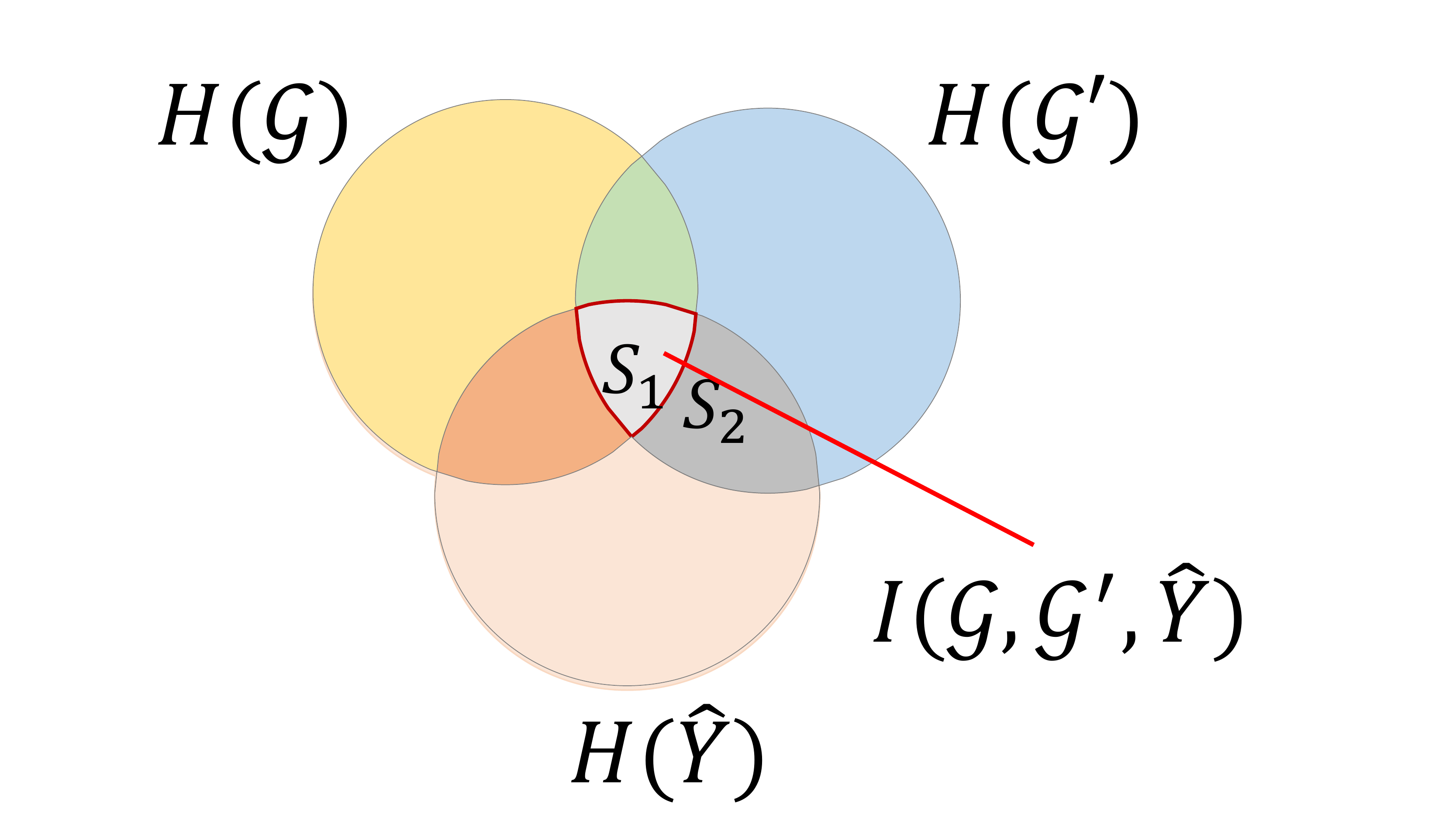}}
    \vskip -1em
    \caption{Illustration of our proposed new objective.} \label{fig:venn}
    \vskip -2em
\end{figure}

From previous discussions, it is shown that $\mathcal{G}'$ obtained via Equation \ref{eq:framework} cannot be safely used as explanations. One main drawback of existing GNN explanation methods lies in the inductive bias that the same outcomes do not guarantee the same causes, leaving existing approaches vulnerable towards spurious explanations.  An illustration is given in Figure~\ref{fig:venn}. Objective proposed in Equation \ref{eq:framework} optimizes the mutual information between explanation candidate $\mathcal{G}'$ and $\hat{Y}$, corresponding to maximize the overlapping between $H(\mathcal{G}')$ and $H(\hat{Y})$ in Figure~\ref{fig:venn_a}, or region $S_1 \cup S_2$ in Figure~\ref{fig:venn_b}. Here, $H$ denotes information entropy. However, this learning target cannot prevent the danger of generating spurious explanations. Provided $\mathcal{G}'$ may fall into the region $S_2$, which cannot faithfully represent graph $\mathcal{G}$. Instead, a more sensible objective should be maximizing region $S_1$ in Figure~\ref{fig:venn_b}. The intuition behind this is that in the search input space that causes the same outcome, identified $\mathcal{G}'$ should account for both representative and discriminative parts of original $\mathcal{G}$, to prevent spurious explanations that produce the same outcomes due to different causes. Concretely, finding $\mathcal{G}'$ that maximize $S_1$ can be formalized as:
\begin{equation}~\label{eq:new_frame}
    \begin{aligned}
    \min_{\mathcal{G}'} &- I(\mathcal{G}, \mathcal{G'}, \hat{Y}), \\
    \quad \text{s.t.} \quad \mathcal{G}' \sim & \mathcal{P}(\mathcal{G}, \mathbf{M}_{A},  \mathbf{M}_{F}) \quad \mathcal{R}(\mathbf{M}_{F},\mathbf{M}_{A}) \leq c 
    \end{aligned}
\end{equation}

\subsection{Connecting to Our Method}
$I(\mathcal{G}, \mathcal{G'}, \hat{Y})$ is intractable as the latent generation mechanism of $\mathcal{G}$ is unknown. In this part, we expand this objective, connect it to Equation~\ref{eq:target}, and construct its proxy optimizable form as: 
\begin{equation} \small
    \begin{aligned}
     I(\mathcal{G}, \mathcal{G'}, \hat{Y}) = & \sum_{y\sim\hat{Y}}\sum_{\mathcal{G}}\sum_{\mathcal{G}'}P(\mathcal{G},\mathcal{G}',y) \cdot \log\frac{P(\mathcal{G}',y)P(\mathcal{G},\mathcal{G}')P(\mathcal{G},y)}{P(\mathcal{G},\mathcal{G}',y)P(\mathcal{G})P(\mathcal{G}')P(y)} \\
    =& \sum_{y\sim\hat{Y}}\sum_{\mathcal{G}}\sum_{\mathcal{G}'}P(\mathcal{G},\mathcal{G}',y) \cdot \log\mathlarger[\frac{P(\mathcal{G}',y)}{P(\mathcal{G}')P(y)}\cdot \frac{P(\mathcal{G},\mathcal{G}')}{P(\mathcal{G})P(\mathcal{G}')} \cdot \frac{P(\mathcal{G},y)}{P(\mathcal{G},y|\mathcal{G}')}\mathlarger] \\
    =& \sum_{y\sim\hat{Y}}\sum_{\mathcal{G}'}P(\mathcal{G}',y)\cdot \log \frac{P(\mathcal{G}',y)}{P(\mathcal{G}')P(y)} +\sum_{\mathcal{G}}\sum_{\mathcal{G}'}P(\mathcal{G},\mathcal{G}')\cdot \log \frac{P(\mathcal{G},\mathcal{G}')}{P(\mathcal{G})P(\mathcal{G}')}\\
    & - \sum_{\mathcal{G}'}\sum_{y\sim\hat{Y}}\sum_{\mathcal{G}}P(\mathcal{G},y,\mathcal{G}')\cdot \log \frac{P(\mathcal{G},y, \mathcal{G}')}{P(\mathcal{G},y)P(\mathcal{G}')}] \\
    =& I(\mathcal{G}', \hat{Y}) + I(\mathcal{G}, \mathcal{G}') - \sum_{y\sim\hat{Y}}\sum_{\mathcal{G}}P(\mathcal{G},y)\sum_{\mathcal{G}'}P(\mathcal{G}'|\mathcal{G},y)\cdot \log P(\mathcal{G}'|\mathcal{G},y) \\
    &+ \sum_{\mathcal{G}'}\sum_{y\sim\hat{Y}}\sum_{\mathcal{G}}P(\mathcal{G},y,\mathcal{G}')\cdot \log P(\mathcal{G}') \\
    =& I(\mathcal{G}', \hat{Y}) + I(\mathcal{G}, \mathcal{G}') + H(\mathcal{G}' | \mathcal{G},\hat{Y})-H(\mathcal{G}').
    \nonumber
    \end{aligned}
\end{equation}
Since both $H(\mathcal{G}' | \mathcal{G},\hat{Y})$ and $H(\mathcal{G}')$ depicts entropy of explanation $\mathcal{G}'$ and are closely related to perturbation budgets, we can neglect these two terms and get a surrogate optimization objective for $\max_{\mathcal{G}'}I(\mathcal{G}, \mathcal{G'}, \hat{Y}) $ as $\max_{\mathcal{G}'}I(\hat{Y},\mathcal{G'})+I(\mathcal{G}',\mathcal{G})$.

In $\max_{\mathcal{G}'}I(\hat{Y},\mathcal{G'})+I(\mathcal{G}',\mathcal{G})$, the first term $\max_{\mathcal{G}'}I(\hat{Y},\mathcal{G'})$ is the same as Eq.(\ref{eq:framework}). Following ~\cite{ying2019gnnexplainer}, We relax it as $\min_{\mathcal{G}'}H_C(\hat{Y}, \hat{Y}'|\mathcal{G'})$, optimizing $\mathcal{G}'$ to preserve original prediction outputs. The second term, $\max_{\mathcal{G}'}I(\mathcal{G}',\mathcal{G})$, corresponds to maximizing consistency between $\mathcal{G}'$ and $\mathcal{G}$. Although the graph generation process is latent, with the safe assumption that embedding $\mathbf{E}_{\mathcal{G}}$ extracted by $f$ is representative of $\mathcal{G}$, we can construct a proxy objective $\max_{\mathcal{G}'}I(\mathbf{E}_{\mathcal{G}'},\mathbf{E}_\mathcal{G})$, improving the consistency in the embedding space. In this work, we optimize this objective by aligning their representations, either optimizing a simplified distance metric or conducting distribution-aware alignment.

%% file: experiment.tex
\section{Experiment}

In this section, we conduct a set of experiments to evaluate the benefits of the proposed auxiliary task in providing instance-level post-hoc explanations. Experiments are conducted on $5$ datasets, and obtained explanations are evaluated with respect to both faithfulness and consistency. Particularly, we aim to answer the following questions:
\begin{itemize}[leftmargin=0.2in]
    \item \textbf{RQ1} Can the proposed framework perform strongly in identifying explanatory sub-structures for interpreting GNNs?
    \item \textbf{RQ2} Is the consistency problem severe in existing GNN explanation methods? Could the proposed embedding alignment improve GNN explainers over this criterion?
    \item \textbf{RQ3} Can our proposed strategies prevent spurious explanations and be more faithful to the target GNN model?
\end{itemize}

\subsection{Experiment Settings}
\subsubsection{Datasets}
We conduct experiments on five publicly available benchmark datasets for explainability of GNNs. The key statistics of the datasets are summarized in Table~\ref{tab:dataset}.
\begin{itemize}[leftmargin=*]
    \item BA-Shapes\cite{ying2019gnnexplainer}: A node classification dataset with a Barabasi-Albert (BA) graph of $300$ nodes as the base structure. $80$ ``house” motifs are randomly attached to the base graph. Nodes in the base graph are labeled as $0$ and those in the motifs are labeled based on positions. Explanations are conducted on those attached nodes, with edges inside the corresponding motif as ground-truth.
    \item Tree-Grid ~\cite{ying2019gnnexplainer}: A node classification dataset created by attaching $80$ grid motifs to a single $8$-layer balanced binary tree. Nodes in the base graph are labeled as $0$ and those in the motifs are labeled as $1$. Edges inside the same motif are used as ground-truth explanations for nodes from class $1$.
    \item Infection~\cite{faber2021comparing}: A single network initialized with an ER random graph. $5\%$ of nodes are labeled as infected, and other nodes are labeled based on their shortest distances to those infected ones. Labels larger than $4$ are clipped. Following~\cite{faber2021comparing}, infected nodes and nodes with multiple shortest paths are neglected. For each node, its shortest path is used as the ground-truth explanation.
    \item Mutag~\cite{ying2019gnnexplainer}: A graph classification dataset. Each graph corresponds to a molecule with nodes for atoms and edges for chemical bonds. Molecules are labeled with consideration of their chemical properties, and discriminative chemical groups are identified using prior domain knowledge. Following PGExplainer~\cite{luo2020parameterized}, chemical groups \textit{$NH_2$} and \textit{$NO_2$} are used as ground-truth explanations.
    \item Graph-SST5~\cite{yuan2020explainability}: A graph classification dataset constructed from text, with labels from sentiment analysis. Each node represents a word and edges denote word dependencies. In this dataset, there is no ground-truth explanation provided, and heuristic metrics are usually adopted for evaluation.
\end{itemize}

\begin{table}[t] \small
  \setlength{\tabcolsep}{4.5pt}
  \caption{Statistics of datasets}\label{tab:dataset} 
  \vskip -1em
  \begin{tabular}{p{1.3cm} | P{1.5cm}  P{1.5cm}   P{1.5cm}  P{1.5cm} P{1.5cm}}

    \hline
     & BA-Shapes &  Tree-Grid  & Infection & Mutag & SST-5 \\
    \hline
    Level & Node & Node & Node &Graph & Graph  \\
    \hline
    Graphs & $1$ & $1$ & $1$ & $4,337$ & $11,855$  \\
    Avg.Nodes & $700$ & $1,231$ & $1,000$ & $30.3$ &  $19.8$\\
    Avg.Edges & $4,110$ & $3,410$ & $4,001$ & $61.5$ & $18.8$ \\
    \hline
    Classes & $4$ & $2$ & $5$ & $2$ & $5$ \\
    \hline
  \end{tabular}
  \vskip -1.5em
\end{table}

\subsubsection{Baselines}
To evaluate the effectiveness of the proposed framework,  we select a group of representative and state-of-the-art instance-level post-hoc GNN explanation methods as baselines. The details are given as follows:
\begin{itemize}[leftmargin=*]
    \item GRAD~\cite{luo2020parameterized}: A gradient-based method, which assigns importance weights to edges by computing gradients of GNN's prediction w.r.t the adjacency matrix.
    \item ATT~\cite{luo2020parameterized}: It utilizes average attention weights inside self-attention layers to distinguish important edges.
    \item GNNExplainer~\cite{ying2019gnnexplainer}: A perturbation-based method which learns an importance matrix separately for every instance.
    \item PGExplaienr~\cite{luo2020parameterized}: A parameterized explainer that learns a GNN to predict important edges for each graph, and is trained via testing different perturbations;
    \item Gem~\cite{lin2021generative}: Similar to PGExplainer but from the causal view, based on the estimated individual causal effect.
    \item RG-Explainer~\cite{shan2021reinforcement}: A reinforcement learning (RL) enhanced explainer for GNN, which constructs $\mathcal{G}'$ by sequentially adding nodes with an RL agent.
\end{itemize}

Our proposed algorithms in Section~\ref{sec:implement} are implemented and incorporated into two representative GNN explanation frameworks, i.e., GNNExplainer~\cite{ying2019gnnexplainer} and PGExplainer~\cite{luo2020parameterized}.

\subsubsection{Configurations}
Following existing work~\cite{luo2020parameterized}, a three-layer GCN~\cite{kipf2016semi} is trained on each dataset as the target model, with the train/validation/test data split as 8:1:1. \revision{The latent dimension is set to $64$ across all datasets, and we use Relu as activation function. } \tianxiang{For graph classification, we concatenate the outputs of global max pooling and global mean pooling as the graph representation.} All explainers are trained using ADAM optimizer with weight decay set to $5e$-$4$. For GNNExplainer, learning rate is initialized to $0.01$ with training epoch being $100$. For PGExplainer, learning rate is initialized to $0.003$ and training epoch is set as $30$. Hyper-parameter $\lambda$, which controls the weight of $\mathcal{L}_{align}$, \revision{is tuned via grid search within the range $[0.1, 10]$ for different datasets separately}. Explanations are tested on all instances. 

\subsubsection{Evaluation Metrics}
To evaluate \textit{faithfulness} of different methods, following~\cite{yuan2020explainability}, we adopt two metrics: (1) AUROC score on edge importance and (2) Fidelity of explanations. On benchmarks with oracle explanations available, we can compute the AUROC score on identified edges as the well-trained target GNN should follow those predefined explanations. On datasets without ground-truth explanations, we evaluate explanation quality with fidelity measurement following~\cite{yuan2020explainability}. Concretely, we observe prediction changes by sequentially removing edges following assigned importance weight, and a faster performance drop represents stronger fidelity. 

To evaluate \textit{consistency} of explanations, we randomly run each method $5$ times, and report average structural hamming distance (SHD)~\cite{tsamardinos2006max} among obtained explanations. A smaller SHD  score indicates stronger consistency.

\subsection{Explanation Faithfulness}
To answer \textbf{RQ1}, we compare explanation methods in terms of AUROC score and explanation fidelity.

\subsubsection{AUROC on Edges}
In this subsection, AUROC scores of different methods are reported by comparing assigned edge importance weight with ground-truth explanations. For baseline methods GRAD, ATT, Gem, and RG-Explainer, their performances reported in their original papers are presented. GNNExplainer and PGExplainer are re-implemented, upon which four alignment strategies are instantiated and tested. Each experiment is conducted $5$ times, and we summarize the average performance in Table~\ref{tab:expl_auroc}. A higher AUROC score indicates more accurate explanations. From the results, we can make the following observations:
\begin{itemize}[leftmargin=*]
    \item Across all four datasets, with both GNNExplainer or PGExplainer as the base method, incorporating embedding alignment can improve the quality of obtained explanations;
    \item \tianxiang{Among proposed alignment strategies, those distribution-aware approaches, particularly the variant based on Gaussian mixture models, achieve the best performance.  In most cases, the variant utilizing latent Gaussian distribution demonstrates stronger improvements, showing the best results on $3$ out of $4$ datasets;
    \item On more complex datasets like Mutag, the benefit of introducing embedding alignment is more significant, e.g., the performance of PGExplainer improves from 83.7\% to 95.9\% with Align\_Gaus. This result also indicates that spurious explanations are severer with increased dataset complexity.   }
\end{itemize}

\subsubsection{Explanation Fidelity}
\begin{figure}[t!]
  \centering
  \subfigure[Tree-Grid, GNNExplainer]{
		\includegraphics[width=0.23\textwidth]{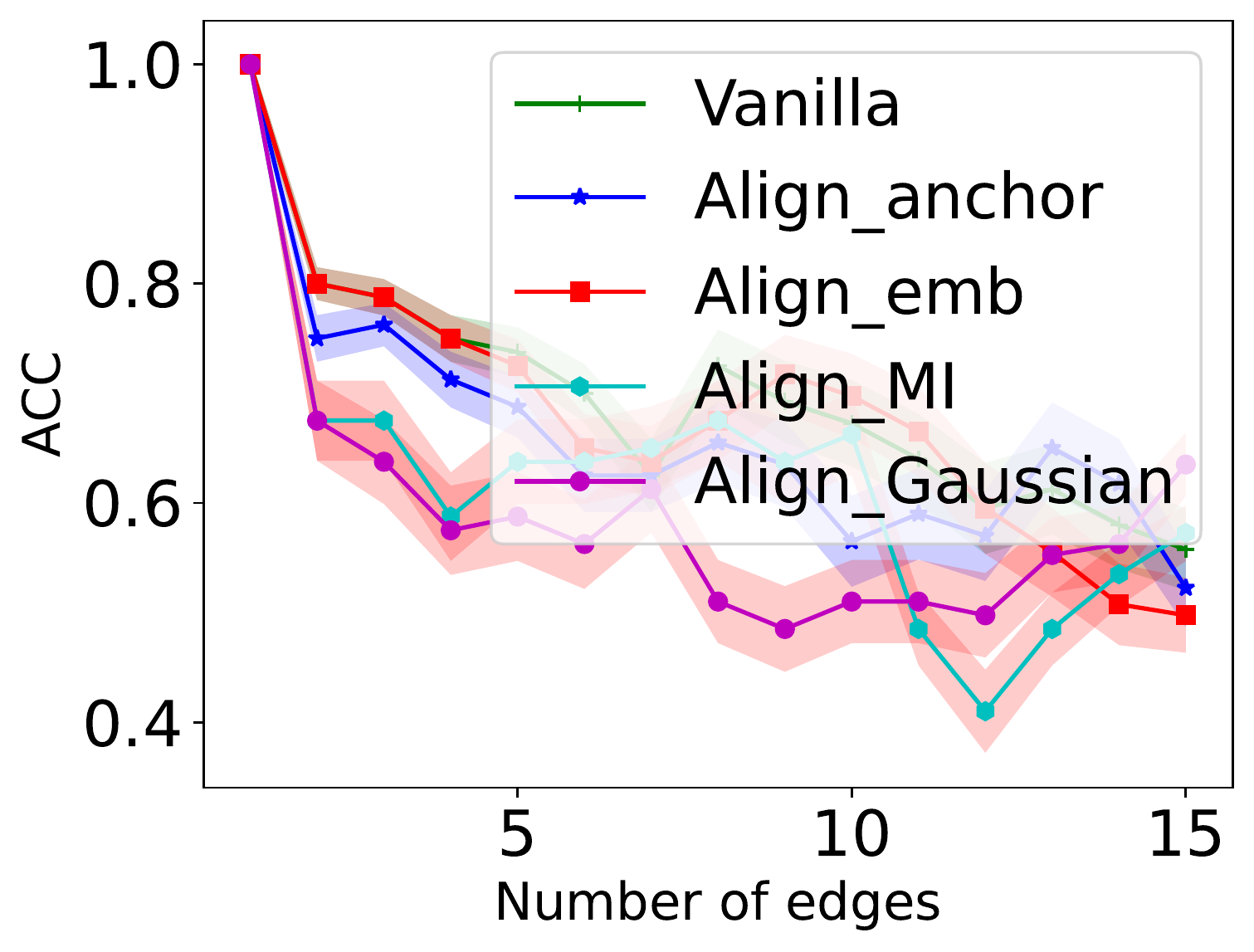}}
  \subfigure[Tree-Grid, PGExplainer]{
		\includegraphics[width=0.23\textwidth]{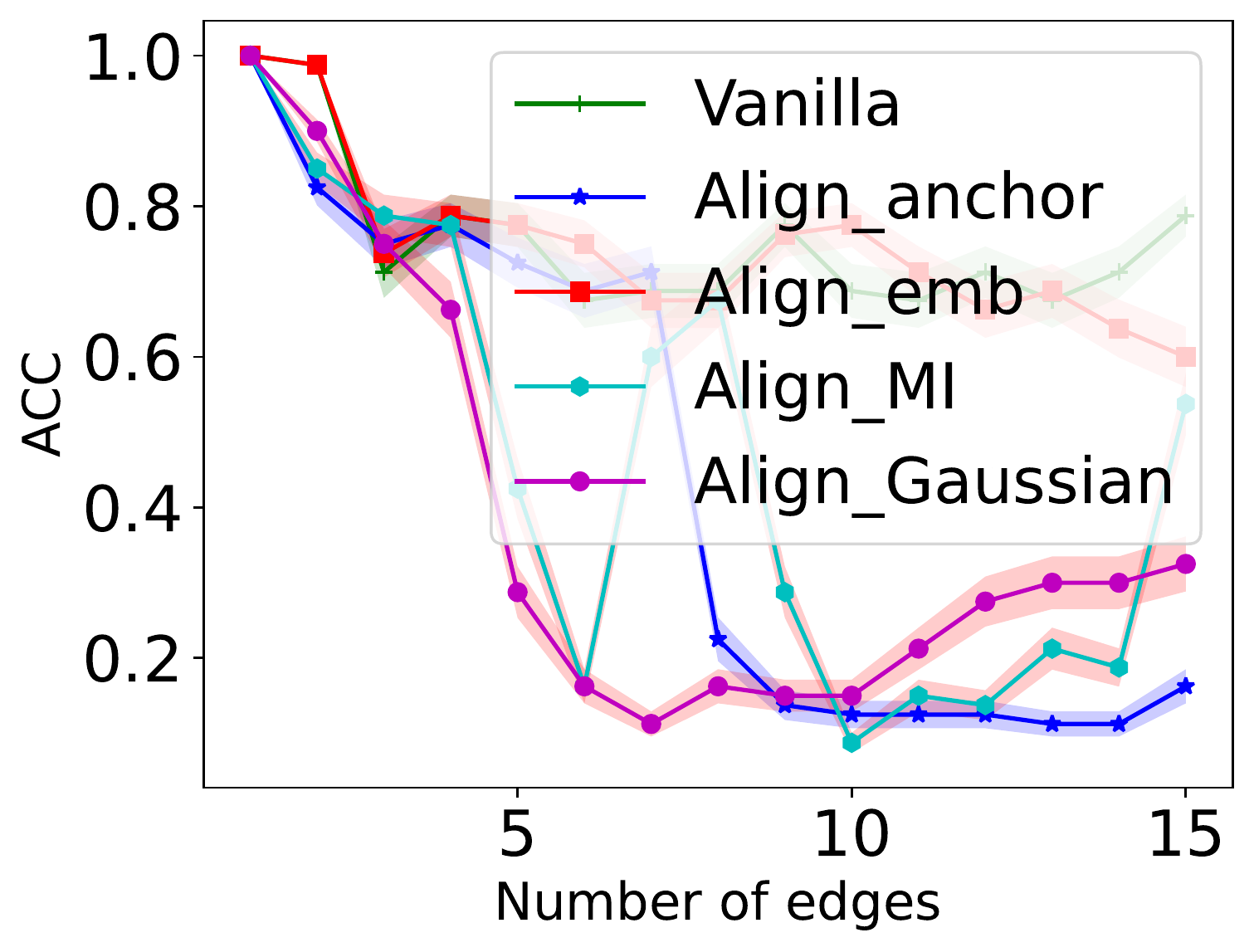}}
  \subfigure[Graph-SS5,  GNNExplainer]{
		\includegraphics[width=0.23\textwidth]{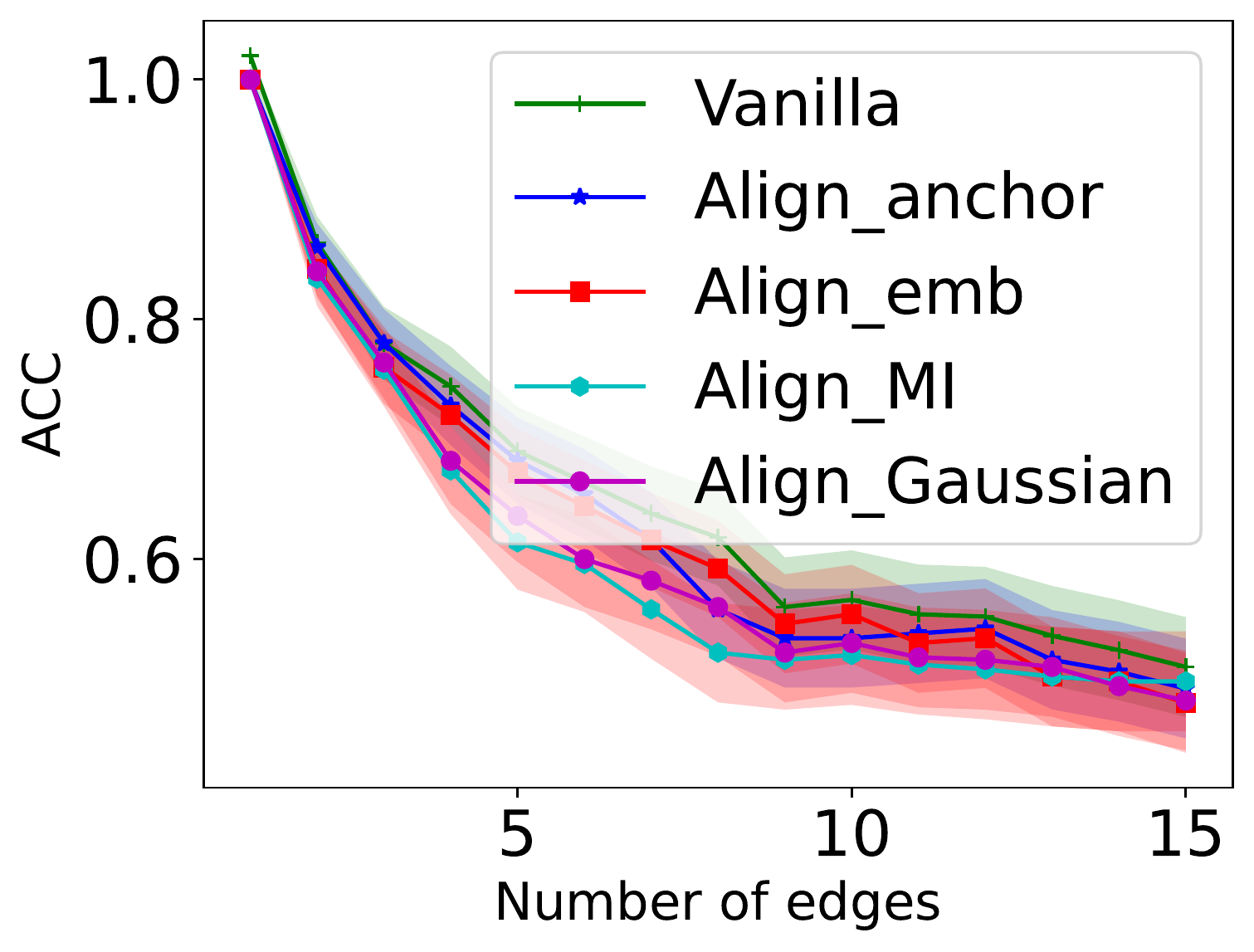}}
  \subfigure[Graph-SS5,  PGExplainer]{
		\includegraphics[width=0.23\textwidth]{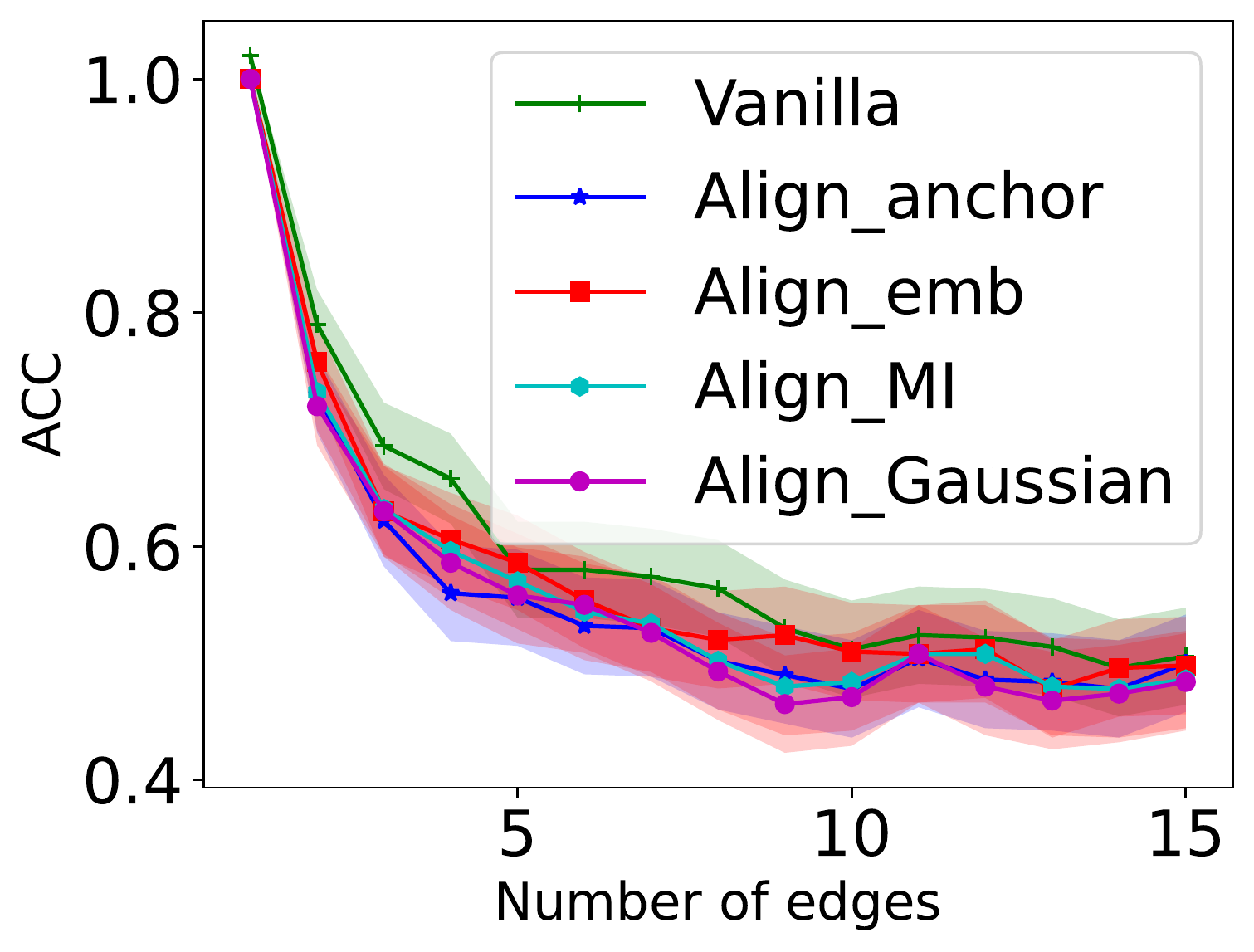}}
    \vskip -1em
    \caption{Explanation Fidelity (Best viewed in color).} \label{fig:fidelity}
    \vskip -2em
\end{figure}

In addition to comparing to ground-truth explanations, we also evaluate the obtained explanations in terms of fidelity. Specifically, we sequentially
remove edges from the graph by following importance weight learned by the explanation model and test the classification performance. Generally, the removal of really important edges would significantly degrade the classification performance. Thus, a faster performance drop represents stronger fidelity. We conduct experiments on Tree-Grid and Graph-SST5. Each experiment is conducted $3$ times and we report results averaged across all instances on each dataset. PGExplainer and GNNExplainer are used as the backbone method. We plot the curves of prediction accuracy concerning the number of removed edges in  Fig.~\ref{fig:fidelity}. From the figure, we can observe that when the proposed embedding alignment is incorporated, the classification accuracy from edge removal drops much faster, which shows that the proposed embedding alignment can help to identify more important edges used by GNN for classification, hence providing better explanations. \tianxiang{Furthermore, distribution-aware alignment strategies like the variant based on Gaussian mixture models demonstrate stronger fidelity in most cases. Besides, it can be noted that on Tree-Grid the fidelity of mutual-information-based alignment is dependent on the number of edges, and achieves better results with edge number within $[8,15]$.}  

\begin{table}[t!]
  \setlength{\tabcolsep}{4.5pt}
  
  \caption{Explanation Faithfulness in terms of AUC on Edges}\label{tab:expl_auroc} 
  \vskip -1em
  \begin{tabular}{p{2.5cm} | P{1.6cm}  P{1.6cm}   P{1.6cm}  P{1.6cm} }

    \hline
     & BA-Shapes &  Tree-Grid  & Infection & Mutag \\
    \hline
    
    \hline
    GRAD & $88.2$ & $61.2$  & $74.0$ & $78.3$ \\
    ATT & $81.5$ & $66.7$  & -- & $76.5$ \\
    Gem & $97.1$ & --  & -- & $83.4$ \\
    RG-Explainer & $98.5$ & $92.7$ & -- & $87.3$ \\
    \hline
    
    \hline
    GNNExplainer & $93.1_{\pm1.8}$ &  $86.2_{\pm2.2}$   & $92.2_{\pm1.1}$ & $74.9_{\pm1.9}$ \\
    + Align\_Emb & $95.3_{\pm1.4}$ &  $91.2_{\pm2.3}$ & $93.0_{\pm1.0}$ & $76.3_{\pm1.7}$ \\
    + Align\_Anchor & $97.1_{\pm1.3}$ & $92.4_{\pm1.9}$ & ${93.1}_{\pm0.8}$ & $78.9_{\pm1.6}$ \\ 
    \hline
    
    + Align\_MI & $97.4_{\pm1.6}$ &  $92.2_{\pm2.5}$ & $93.2_{\pm1.0}$ & $78.2_{\pm1.8}$ \\
    + Align\_Gaus & $96.7_{\pm1.3}$ & $92.5_{\pm1.9}$ & ${93.1}_{\pm0.9}$ & $79.3_{\pm1.5}$ \\    
    \hline
    
    \hline
    PGExplainer & $96.9_{\pm0.7}$ & $92.7_{\pm1.5}$ &   $89.6_{\pm0.6}$ & $83.7_{\pm1.2}$ \\
    + Align\_Emb & $97.2_{\pm0.7}$ & ${95.8}_{\pm0.9}$ &   $90.5_{\pm0.7}$ & $92.8_{\pm1.1}$ \\ 
    + Align\_Anchor & ${98.7}_{\pm0.5}$ & $94.7_{\pm1.2}$   & $91.6_{\pm0.6}$ & ${94.5}_{\pm0.8}$ \\
    \hline
    + Align\_MI & $\mathbf{99.3}_{\pm0.8}$ &  $96.2_{\pm1.3}$ & $92.0_{\pm0.4}$ & $94.3_{\pm1.1}$ \\
    + Align\_Gaus & $99.2_{\pm0.3}$ & $\mathbf{96.4}_{\pm1.1}$ & $\mathbf{92.5}_{\pm0.8}$ & $\mathbf{95.9}_{\pm1.2}$ \\  
    \hline

    \hline
  \end{tabular}
  \vskip -1em
\end{table}

From these two experiments, we can observe that embedding alignment can obtain explanations of better faithfulness and is flexible to be incorporated into various models such as GNNExplainer and PGExplainer, which answers RQ1.

\subsection{Explanation Consistency}
One problem of spurious explanation is that, due to the randomness in initialization of the explainer, the explanation for the same instance given by a GNN explainer could be different for different runs, which violates the \textit{consistency} of explanations. To test the severity of this problem and answer \textbf{RQ2}, we evaluate the proposed framework in terms of explanation consistency. We adopt GNNExplainer and PGExplainer as baselines. Specifically, SHD distance among explanatory edges with top-$k$ importance weights identified each time is computed. Then, the results are averaged for all instances in the test set. Each experiment is conducted $5$ times, and the average consistency on dataset Tree-Grid and Mutag are reported in Table~\ref{tab:consistency_Tree} and Table~\ref{tab:consistency_Mutag}, respectively. Larger distances indicate inconsistent explanations. From the table, we can observe that existing method following Equation~\ref{eq:framework} suffers from the consistency problem. For example, average SHD distance on top-$6$ edges is $4.39$ for GNNExplainer. Introducing the auxiliary task of aligning embeddings can significantly improve explainers in terms of this criterion. \tianxiang{Of these different alignment strategies, the variant based on Gaussian mixture model shows the strongest performance in most cases. After incorporating Align\_Gaus on dataset TreeGrid, SHD distance of top-$6$ edges drops from $4.39$ to $2.13$ for GNNExplainer and from $1.38$ to $0.13$ for PGExplainer. After incorporating it on dataset Mutag, SHD distance of top-$6$ edges drops from $4.78$ to $3.85$ for GNNExplainer and from $3.42$ to $1.15$ for PGExplainer. } These results validate the effectiveness of our proposal in obtaining consistent explanations.

\begin{table}[t!] \centering \small
  \setlength{\tabcolsep}{4.5pt}
  
  \caption{Consistency of explanation in terms of average SHD across $5$ rounds of random running on Tree-Grid.}\label{tab:consistency_Tree} 
  \vskip -1em
  \begin{tabular}{p{2.2cm} |  P{0.9cm}  P{0.9cm}  P{0.9cm}  P{0.9cm}  P{0.9cm}  P{0.9cm}}

    \hline
     &  \multicolumn{6}{c}{Top-K Edges}  \\
    \hline

    \hline
    Methods & 1 & 2 & 3 & 4 & 5 & 6  \\
    \hline

    \hline
    GNNExplainer & $0.86$ & $1.85$ & $2.48$ & $3.14$ & $3.77$ & $4.39$ \\
    +Align\_Emb & $0.77$ & $1.23$ & $1.28$ & $0.96$ & $1.81$ & $2.72$ \\
     +Align\_Anchor & $0.72$ & $1.06$ & $0.99$ & $0.53$ & $1.52$ & $2.21$ \\
     \hline
     +Align\_MI  & $0.74$ & $1.11$ & $1.08$ & $1.32$ & $1.69$ & $2.27$  \\
     +Align\_Gaus  & $0.68$ & $1.16$ & $1.13$ & $0.72$ & $1.39$ & $2.13$ \\
    \hline

    \hline
    PGExplainer  & $0.74$ & $1.23$ & $0.76$ & $0.46$ & $0.78$ & $1.38$ \\
    +Align\_Emb  & $0.11$ & $0.15$ & ${0.13}$ & $\textbf{0.11}$ & $0.24$ & $0.19$ \\
    +Align\_Anchor  & ${0.07}$ & ${0.12}$ & ${0.13}$ & $0.16$ & ${0.21}$ & $\textbf{0.13}$  \\
    \hline
    +Align\_MI & $0.28$ & $0.19$ & $0.27$ & $0.15$ & $0.20$ & $0.16$ \\
     +Align\_Gaus & $\textbf{0.05}$ & $\textbf{0.08}$ & $\textbf{0.10}$ & $0.12$ & $\textbf{0.19}$ & $\textbf{0.13}$ \\
    \hline

    \hline
  \end{tabular}
\end{table}

\begin{table}[t!] \centering \small
  \setlength{\tabcolsep}{4.5pt}
  
  \caption{Consistency of explanation in terms of average SHD distance across $5$ rounds of random running on Mutag.}\label{tab:consistency_Mutag} 
  \vskip -1em
  \begin{tabular}{p{2.2cm} |  P{0.9cm}  P{0.9cm}  P{0.9cm}  P{0.9cm}  P{0.9cm}  P{0.9cm}}

    \hline
     &  \multicolumn{6}{c}{Top-K Edges}  \\
    \hline

    \hline
    Methods & 1 & 2 & 3 & 4 & 5 & 6  \\
    \hline

    \hline
    GNNExplainer & $1.12$ & $1.74$ & $2.65$ & $3.40$ & $4.05$ & $4.78$ \\
    +Align\_Emb & $1.05$ & $1.61$ & $2.33$ & $3.15$ & $3.77$ & $4.12$ \\
     +Align\_Anchor & $1.06$ & $1.59$ & $2.17$ & $3.06$ & $3.54$ & $3.95$ \\
     \hline
    +Align\_MI & $1.11$ & $1.68$ & $2.42$ & $3.23$ & $3.96$ & $4.37$ \\
     +Align\_Gaus & $1.03$ & $1.51$ & $2.19$ & $3.02$ & $3.38$ & $3.85$ \\
    \hline

    \hline
    PGExplainer  & $0.91$ & $1.53$ & $2.10$ & $2.57$ & $3.05$ & $3.42$ \\
    +Align\_Emb  & $0.55$ & $0.96$ & $1.13$ & $1.31$ & $1.79$ & $2.04$ \\
    +Align\_Anchor  & $\textbf{0.51}$ & $\textbf{0.90}$ & $\textbf{1.05}$ & ${1.27}$ & ${1.62}$ & ${1.86}$  \\
    \hline
    +Align\_MI & $0.95$ & $1.21$ & $1.73$ & $2.25$ & $2.67$ & $2.23$ \\
     +Align\_Gaus & $0.59$ & $1.34$ & $1.13$ & $\textbf{0.84}$ & $\textbf{1.25}$ & $\textbf{1.15}$ \\
    \hline

    \hline
  \end{tabular}
    \vskip -1em
\end{table}

\subsection{Ability in Avoiding Spurious Explanations}
Existing graph explanation benchmarks are usually designed to be less ambiguous, containing only one oracle cause of labels, and identified explanatory substructures are evaluated via comparing with the ground-truth explanation. However, this result could be misleading, as faithfulness of explanation in more complex scenarios is left untested. Real-world datasets are usually rich in spurious patterns and a trained GNN could contain diverse biases, setting a tighter requirement on explanation methods. Thus, to evaluate if our framework can alleviate the spurious explanation issue and answer \textbf{RQ3}, we create a new graph-classification dataset: MixMotif, which enables us to train a biased GNN model, and test whether explanation methods can successfully expose this bias. 

Specifically, inspired by ~\cite{wu2022discovering}, we design three types of base graphs, i.e., Tree, Ladder, and Wheel, and three types of motifs, i.e., Cycle, House, and Grid. With a mix ratio $\gamma$, motifs are preferably attached to base graphs. For example, Cycle is attached to Tree with probability $\frac{2}{3}\gamma+\frac{1}{3}$, and to others with probability $\frac{1-\gamma}{3}$. So are the cases for House to Ladder and Grid to Wheel. Labels of obtained graphs are set as type of the motif. When $\gamma$ is set to $0$, each motif has the same probability of being attached to the three base graphs. In other words, there's no bias on which type of base graph to attach for each type of motif. Thus, we consider the dataset with $\gamma=0$ as clean or bias-free. We would expect GNN trained on data with $\gamma=0$ to focus on the motif structure for motif classification. However, when $\gamma$ becomes larger, the spurious correlation between base graph and the label would exist, i.e., a GNN might utilize the base graph structure for motif classification instead of relying on the motif structure. For each setting, the created dataset contains $3,000$ graphs, and train:evaluation:test are split as $5:2:3$.

In this experiment, we set $\gamma$ to $0$ and $0.7$ separately, and train GNN $f_{0}$ and $f_{0.7}$ for each setting. Two models are tested in graph classification performance. Then, explanation methods are applied to and fine-tuned on $f_0$. Following that, these explanation methods are applied to explain $f_{0.7}$ using found hyper-parameters. Results are summarized in Table~\ref{tab:biasedExpl}.

\begin{table}[t!]
  \setlength{\tabcolsep}{4.5pt}
  
  \caption{Performance on MixMotif. Two GNNs are trained with different $\gamma$. We check their performance in graph classification, then compare obtained explanations with the motif. }
  \vskip -1em
  \begin{tabular}{p{1.2cm} | p{0.5cm} |  P{1.4cm}  P{1.4cm} | P{1.4cm}  P{1.4cm} }

    \hline
    \multicolumn{2}{c}{} &  \multicolumn{4}{c}{$\gamma$ in Training}  \\
    \hline
    \multicolumn{2}{c|}{Classification} & \multicolumn{2}{c|}{$0$} & \multicolumn{2}{c}{$0.7$} \\
    \hline
    \multirow{2}{*}{$\gamma$ in test} & $0$ & \multicolumn{2}{c|}{$0.982$}  &  \multicolumn{2}{c}{$0.765$} \\
     &$0.7$ & \multicolumn{2}{c|}{$0.978$}  &  \multicolumn{2}{c}{$0.994$} \\
    \hline 
    \multicolumn{2}{c|}{Explanation} & PGExplainer & +Align & PGExplainer & +Align \\
    \hline
    \multicolumn{2}{c|}{\multirow{2}{*}{\parbox{1.4cm}{AUROC on Motif}}} & $0.711$ & $\mathbf{0.795}$ & $0.748$ & $\mathbf{0.266}$ \\
    \multicolumn{2}{c|}{}& \multicolumn{2}{c|}{(Higher is better)} & \multicolumn{2}{c}{(Lower is better)} \\
    \hline
  \end{tabular}\label{tab:biasedExpl}
  \vskip -1.5em
\end{table}

From Table~\ref{tab:biasedExpl}, we can observe that (1) $f_0$ achieves almost perfect graph classification performance during testing. This high accuracy indicates that it captures the genuine pattern, relying on motifs to make predictions. Looking at explanation results, it is shown that our proposal offers more faithful explanations, achieving higher AUROC on motifs. (2) $f_{0.7}$ fails to predict well with $\gamma=0$, showing that there are biases in it and it no longer depends solely on the motif structure for prediction. Although ground-truth explanations are unknown in this case, a successful explanation should expose this bias. However, PGExplainer would produce similar explanations as the clean model, still highly in accord with motif structures. Instead, for explanations produced by embedding alignment, AUROC score would drop from $0.795$ to $0.266$, exposing the change in prediction rationales, hence able to expose biases. (3) In summary, our proposal can provide more faithful explanations for both clean and mixed settings, while PGExplainer would suffer from spurious explanations and fail to faithfully explain GNN's predictions, especially in the existence of biases.


\subsection{Hyperparameter Sensitivity Analysis}
\begin{figure}[t!]
  \centering
  \subfigure[Tree-Grid]{
		\includegraphics[width=0.35\textwidth]{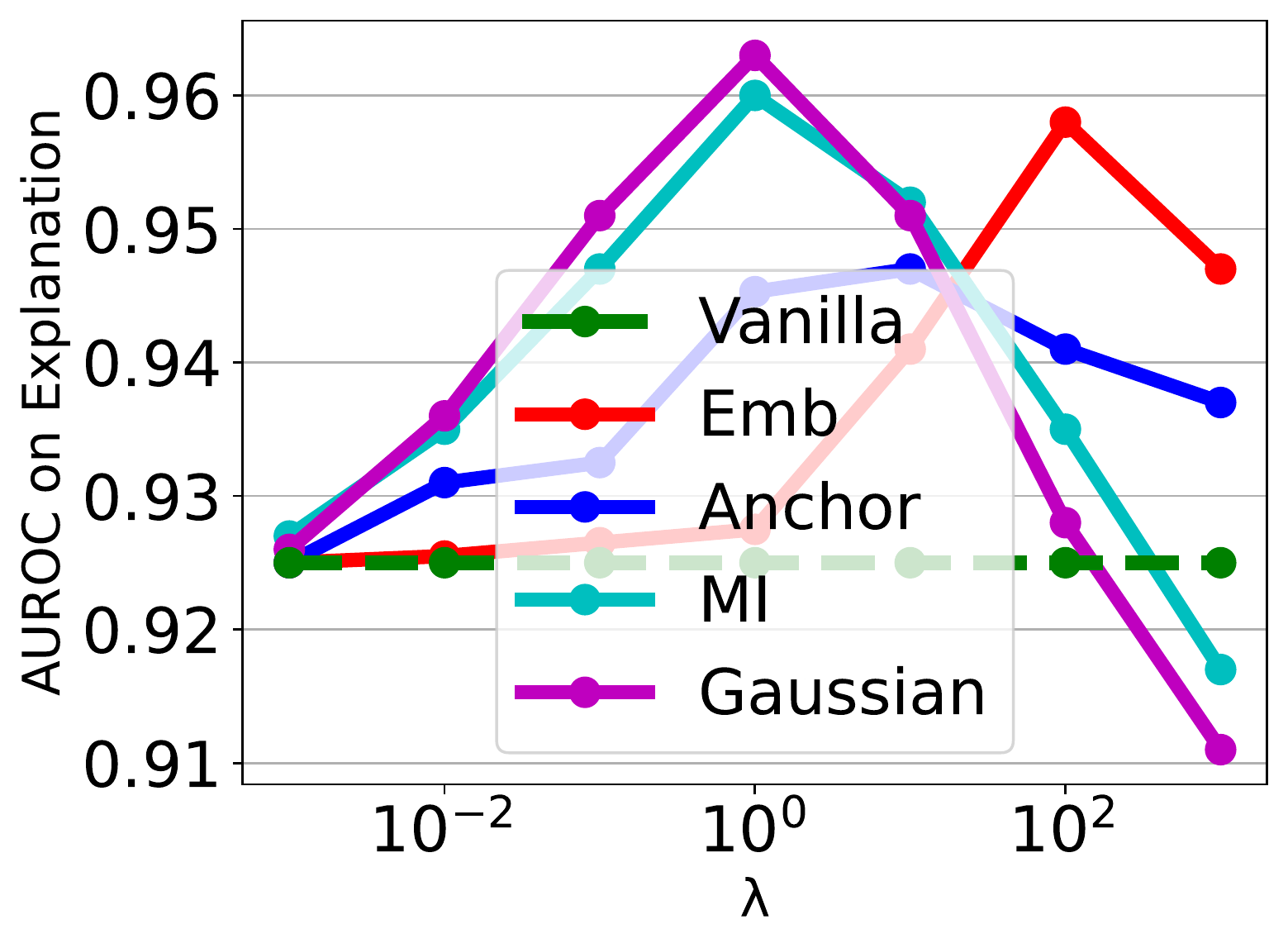}}
  \subfigure[Mutag]{
		\includegraphics[width=0.355\textwidth]{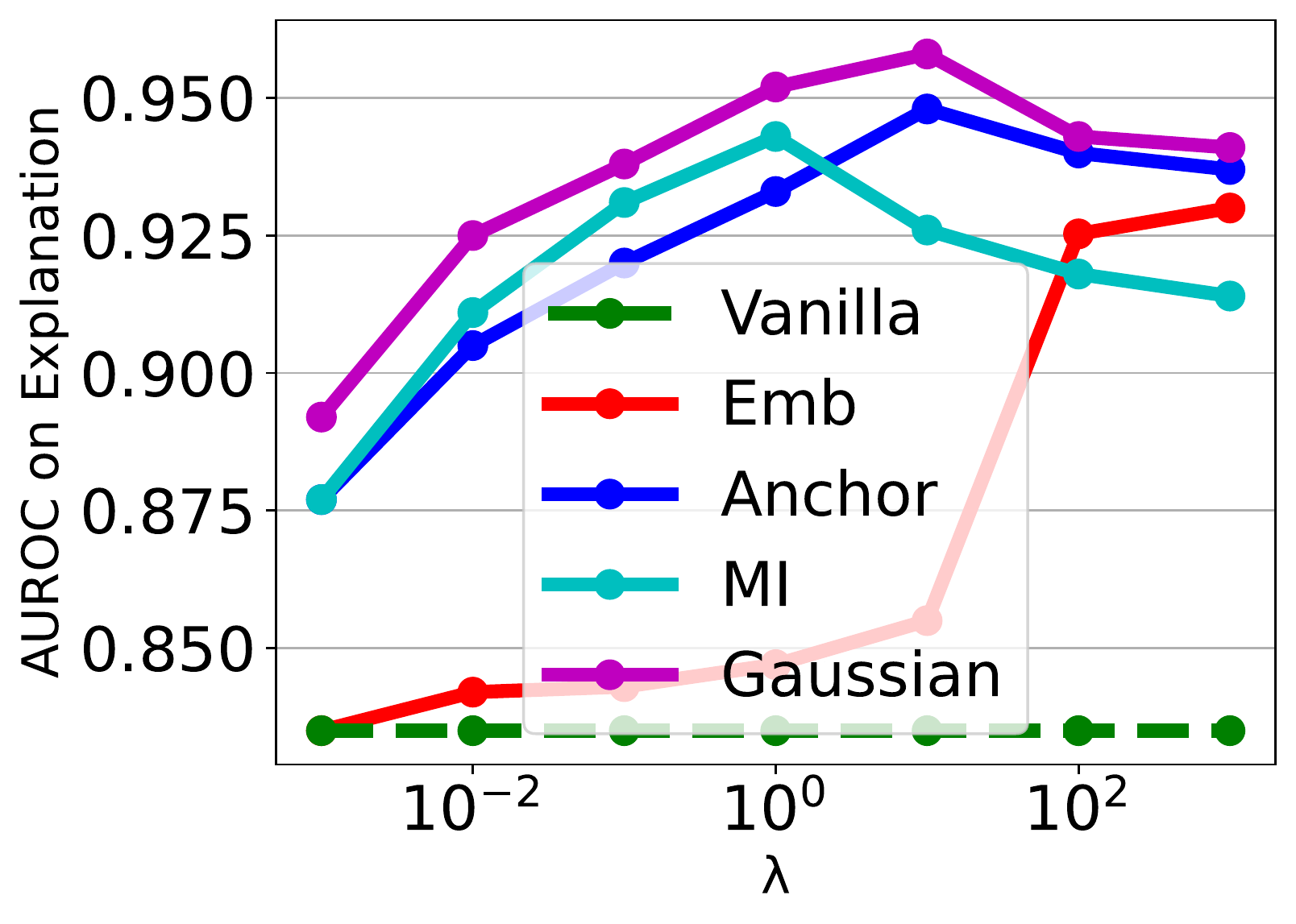}}
    \vskip -1em
    \caption{Sensitivity of PGExplainer towards weight of embedding alignment loss. } \label{fig:parameter_sensitivity}
    \vskip -1em
\end{figure}

In this part, we vary the hyper-parameter $\lambda$ to test the sensitivity of the proposed framework toward its values. $\lambda$ controls the weight of our proposed embedding alignment task. To keep simplicity, all other configurations are kept unchanged, and $\lambda$ is varied within the scale $[1e-3,1e-2, 1e-1, 1, 10,1e2, 1e3\}$. PGExplainer is adopted as the base method. Experiments are randomly conducted $3$ times on dataset Tree-Grid and Mutag. Averaged results are visualized in Figure~\ref{fig:parameter_sensitivity}. From the figure, we can make the following observations:
\begin{itemize}[leftmargin=*]
    \item For all four variants, increasing $\lambda$ has a positive effect at first, and the further increase would result in a performance drop. For example on the Tree-Grid dataset, best results of variants based on anchors, latent Gaussian mixture models and mutual information scores are all obtained with $\lambda$ around $1$. When $\lambda$ is small, the explanation alignment regularization in Eq.~\ref{eq:target} will be underweighted. On the other hand, a too-large $\lambda$ may underweight the MMI-based explanation framework, which preserves the predictive power of obtained explanations. 
    \item Among these variants, the strategy based on latent Gaussian mixture models shows the strongest performance in most cases. For example, for both datasets Tree-Grid and Mutag, this variant achieves the highest AUROC scores on identified explanatory edges. On the other hand, the variant directly using Euclidean distances shows inferior performances in most cases. We attribute this to their different ability in modeling the distribution and conducting alignment.
\end{itemize}

%% file: conclusion.tex
\section{Conclusion} \label{sec:conclusion}
In this work, we study a novel problem of obtaining faithful and consistent explanations for GNNs, which is largely neglected by existing MMI-based explanation framework. With close analysis on the inference of GNNs, we propose a simple yet effective approach by aligning internal embeddings. Theoretical analysis shows that it is more faithful in design, optimizing an objective that encourages high MI between the original graph, GNN output, and identified explanation. Four different strategies are designed, by directly adopting Euclidean distance, using anchors, KL divergence with Gaussian mixture models, and estimated MI scores. All these algorithms can be incorporated into existing methods with no effort. Experiments validate their effectiveness in promoting the faithfulness and consistency of explanations. 

In the future, we will seek more robust explanations. Increased robustness indicates stronger generality, and could provide better class-level interpretation at the same time. 
Besides, the evaluation of explanation methods also needs further studies. Existing benchmarks are usually clear and unambiguous, failing to simulate complex real-world scenarios.